%% file: main.tex
\documentclass[10pt,a4paper]{main}

\usepackage[all]{hypcap}
\usepackage[authoryear, round]{natbib}
\usepackage{nicefrac}
\usepackage{multirow}
\usepackage[noabbrev,capitalise]{cleveref}
\usepackage{dsfont}
\usepackage{makecell}
\usepackage{subcaption}
\usepackage{bbm}
\usepackage{wrapfig}
\usepackage{mathtools,xparse}
\usepackage{array}
\usepackage{arydshln}
\usepackage{scalerel}
\usepackage{tablefootnote}
\usepackage{tocloft}
\usepackage{adjustbox}
\usepackage{setspace}
\usepackage{fontawesome5}
\usepackage{xstring}
\usepackage{thmtools,thm-restate}
\usepackage[most,skins,theorems]{tcolorbox}

\newtcolorbox{promptbox}{
  colback=gray!5,
  colframe=gray!70,
  boxrule=0.8pt,
  arc=2mm,
  left=10pt,
  right=10pt,
  top=10pt,
  bottom=10pt,
  width=\linewidth,
  fontupper=\ttfamily\footnotesize,
  enhanced,
}

\newboolean{showsection}
\setboolean{showsection}{true}

\definecolor{custom_green}{rgb}{0.0, 0.5, 0.0}
\definecolor{custom_red}{rgb}{1.0, 0.01, 0.24}
\definecolor{custom_blue}{HTML}{C9DAF7}
\definecolor{custom_purple}{HTML}{D9D1E9}
\definecolor{title_blue}{HTML}{204899}
\definecolor{cite_blue}{HTML}{044dc1}
\definecolor{cite_purple}{HTML}{7406a7}
\newcommand{\redx}{\textcolor{custom_red}{\ding{55}}}
\newcommand{\greencheck}{\textcolor{custom_green}{\ding{51}}}

\hypersetup{
  colorlinks = true,
  citecolor  = {cite_blue},
  linkcolor  = {cite_purple},
  urlcolor   = {cite_purple},
}

\input{macro}

\input{math_commands}

\input{frame_style}

\makeatletter
\def\mathcolor#1#{\@mathcolor{#1}}
\def\@mathcolor#1#2#3{%
  \protect\leavevmode
  \begingroup
    \color#1{#2}#3%
  \endgroup
}
\makeatother

\crefformat{equation}{#2Eq.\;(#1)#3}
\Crefformat{equation}{#2Eq.\;(#1)#3}
\crefname{assumption}{Assumption}{Assumptions}
\Crefname{assumption}{Assumption}{Assumptions}

\usepackage{crossreftools}
\title{UniversalRAG: Retrieval-Augmented Generation\\over Corpora of Diverse Modalities and Granularities}
\displaytitle{UniversalRAG: Retrieval-Augmented Generation over Corpora of Diverse Modalities and Granularities}

\reportnumber{}

\author[1\footnotesize{\textbf{*}}]{Woongyeong Yeo}
\author[1\footnotesize{\textbf{*}}]{Kangsan Kim}
\author[1]{Soyeong Jeong}
\author[1$\dagger$]{Jinheon Baek}
\author[1,2$\dagger$]{Sung Ju Hwang}

\affil[1]{KAIST}
\affil[2]{DeepAuto.ai}

\paperurl{https://universalrag.github.io}
\authoremails{\{wgcyeo, kangsan.kim, starsuzi, jinheon.baek, sungju.hwang\}@kaist.ac.kr}

\input{sections/0_abstract}

\begin{document}

\maketitle

{
    \renewcommand{\thefootnote}{\fnsymbol{footnote}}
    \footnotetext[1]{Equal contribution; \textsuperscript{$\dagger$}Equal advising.}
}

\input{sections/1_introduction}
\input{sections/2_method}
\input{sections/3_experiment}
\input{sections/4_related_work}
\input{sections/5_conclusion}

\input{sections/x_limitation}
\input{sections/x_ethics}
\input{sections/x_acknowledgement}

\bibliographystyle{plainnat}
\bibliography{main}

\clearpage
\appendix{\input{sections/x_appendix}}

\end{document}

%% file: macro.tex
\definecolor{blanchedalmond}{rgb}{1.0, 0.92, 0.8}
\definecolor{carmine}{rgb}{0.59, 0.0, 0.09}
\definecolor{lightblue}{rgb}{0.22,0.45,0.70}%

\newtheorem{theorem}{Theorem}[section]

\newtheorem{proposition}[theorem]{Proposition}

\newtheorem{remark}[theorem]{Remark}

\renewcommand{\mathbf}{\boldsymbol}

\makeatletter
\def\Ddots{\mathinner{\mkern1mu\raise\p@
\vbox{\kern7\p@\hbox{.}}\mkern2mu
\raise4\p@\hbox{.}\mkern2mu\raise7\p@\hbox{.}\mkern1mu}}
\makeatother

\numberwithin{equation}{section}

\definecolor{amaranth}{rgb}{0.9, 0.17, 0.31}
\definecolor{antiquebrass}{rgb}{0.8, 0.58, 0.46}
\definecolor{antiquefuchsia}{rgb}{0.57, 0.36, 0.51}
\definecolor{chromeyellow}{rgb}{0.31, 0.47, 0.26}


%% file: math_commands.tex
\usepackage{amsmath,amsfonts,bm}

\def\eqref#1{equation~\ref{#1}}

\def\1{\bm{1}}

\def\va{{\bm{a}}}

\def\vc{{\bm{c}}}

\def\vo{{\bm{o}}}

\def\vq{{\bm{q}}}
\def\vr{{\bm{r}}}
\def\vs{{\bm{s}}}

\def\vx{{\bm{x}}}
\def\vy{{\bm{y}}}

\DeclareMathAlphabet{\mathsfit}{\encodingdefault}{\sfdefault}{m}{sl}
\SetMathAlphabet{\mathsfit}{bold}{\encodingdefault}{\sfdefault}{bx}{n}



%% file: frame_style.tex
\newcommand{\whitebox}{%
  \hfill\textcolor{white}{\rule[0.8mm]{1.6mm}{1.6mm}}\hfill%
}

\newcommand{\filmbox}[1]{%
  \IfBeginWith{#1}{'}{%
    \StrGobbleLeft{#1}{1}[\@temp]
    \StrGobbleRight{\@temp}{1}[\filename]%
  }{%
    \edef\filename{#1}%
  }%
  \setlength{\fboxsep}{0pt}%
  \colorbox{black}{%
    \begin{minipage}{2.5cm}
      \rule{0mm}{3.2mm}%
      \whitebox\whitebox\whitebox\whitebox\whitebox%
      \whitebox\whitebox\whitebox\whitebox\null\\
      \null\hfill
        \includegraphics[width=2.4cm]{\filename}%
      \hfill\null\\[0mm]
      \null
      \whitebox\whitebox\whitebox\whitebox\whitebox%
      \whitebox\whitebox\whitebox\whitebox\null
    \end{minipage}%
  }%
}

\newcommand{\whiteboxsmall}{%
  \hfill\textcolor{white}{\rule[0.4mm]{0.8mm}{0.8mm}}\hfill%
}

\newcommand{\filmboxsmall}[1]{%
  \IfBeginWith{#1}{'}{%
    \StrGobbleLeft{#1}{1}[\@temp]
    \StrGobbleRight{\@temp}{1}[\filename]%
  }{%
    \edef\filename{#1}%
  }%
  \setlength{\fboxsep}{0pt}%
  \colorbox{black}{%
    \begin{minipage}{1.5cm}
      \rule{0mm}{1.6mm}%
      \whiteboxsmall\whiteboxsmall\whiteboxsmall\whiteboxsmall\whiteboxsmall%
      \whiteboxsmall\whiteboxsmall\whiteboxsmall\whiteboxsmall\null\\
      \null\hfill
        \includegraphics[width=1.4cm]{\filename}%
      \hfill\null\\[-0.9mm]
      \null
      \whiteboxsmall\whiteboxsmall\whiteboxsmall\whiteboxsmall\whiteboxsmall%
      \whiteboxsmall\whiteboxsmall\whiteboxsmall\whiteboxsmall\null
    \end{minipage}%
  }%
}

%% file: sections/0_abstract.tex
\begin{abstract}

Retrieval-Augmented Generation (RAG) has shown substantial promise in improving factual accuracy by grounding model responses with external knowledge relevant to queries. However, most existing approaches are limited to a text-only corpus, and while recent efforts have extended RAG to other modalities such as images and videos, they typically operate over a single modality-specific corpus. In contrast, real-world queries vary widely in the type of knowledge they require, which a single type of knowledge source cannot address. To address this, we introduce UniversalRAG, an any-to-any RAG framework designed to retrieve and integrate knowledge from heterogeneous sources with diverse modalities and granularities. Specifically, motivated by the observation that forcing all modalities into a unified representation space derived from a single aggregated corpus causes a modality gap, where the retrieval tends to favor items from the same modality as the query, we propose modality-aware routing, which dynamically identifies the most appropriate modality-specific corpus and performs targeted retrieval within it, and further justify its effectiveness with a theoretical analysis. Moreover, beyond modality, we organize each modality into multiple granularity levels, enabling fine-tuned retrieval tailored to the complexity and scope of the query. We validate UniversalRAG on 10 benchmarks of multiple modalities, showing its superiority over various modality-specific and unified baselines.

\end{abstract}

%% file: sections/1_introduction.tex
\input{figures/fig_concept}

\section{Introduction}

Large Language Models (LLMs) have demonstrated remarkable performance across various tasks, and have been widely adopted to assist users in everyday life~\citep{Gemini, gpt5}. However, LLMs often generate factually incorrect or misleading information, especially on topics they were less or not exposed to during training~\citep{SirenSurvey, HallucinationSurvey}. To address this, Retrieval-Augmented Generation (RAG) has emerged as a promising approach, which allows the model responses to be grounded in the query-relevant knowledge retrieved from external knowledge sources, enhancing factual accuracy~\citep{RAG, RAGSurvey, BenchmarkLLMRAG}.

Despite its effectiveness, existing approaches are typically designed for a single corpus and modality, limiting their ability to address queries that require diverse knowledge sources. In practice, as shown in \cref{fig:fig_concept}, user queries vary widely in the type of knowledge they require: some are best answered using text (e.g., surface-level facts), others demand visual understanding from images or videos (spatial or temporal cues), and still others require combinations of these modalities. Yet, the field of RAG primarily originates with a textual corpus~\citep{RAG, ActiveRAG, CorrectiveRAG}, and although recent efforts have expanded it to modalities beyond text (such as images and videos)~\citep{AskAnyModality, VideoRAG, ImageRAG}, existing RAG methods (when considered individually) are typically modality- and corpus-specific; therefore, they may be suboptimal to serve as a universal, one-for-all framework that can flexibly handle the wide range of queries, whose knowledge requirements vary.

\input{figures/fig_embedding}
In this work, we present UniversalRAG, a novel any-to-any RAG framework that brings together knowledge distributed across multiple modality-specific corpora, and leverages them to generate grounded responses to queries in a universal workflow. To operationalize this, one straightforward approach might be to aggregate all entries from the collected, heterogeneous knowledge corpora, and embed them into a unified space using a multimodal encoder (which is typically trained to align inputs from different modalities if they are semantically similar). However, despite such alignment efforts, we find that this strategy suffers from modality gaps~\citep{modalitygap, vlm2vecv2}, the tendency that inputs are clustered based on their modality rather than their semantic relevance (visualized in \cref{fig:embedding,fig:appendix_embedding}). As a result, retrieval becomes biased toward knowledge sources that share the same modality as the query, overlooking relevant content from other modalities.

To address this challenge, rather than forcing all modalities into a single embedding space, we take a different direction and introduce \emph{modality-aware routing}. UniversalRAG predicts its modality requirements and routes retrieval to the corresponding modality-specific corpora (potentially multiple, when the query calls for cross-modal evidence), after which the retrieved knowledge is jointly used for grounding. Notably, this strategy not only sidesteps modality gaps by avoiding every cross-modal comparison but also enables seamless integration of new modalities by extending the routing logic without modifying existing modality-specific retrievers.

Beyond modality, data granularity (i.e., the size or unit of each entry in the corpus) also affects retrieval precision and generation quality~\citep{DenseXRetireval, MoG}, since different queries benefit from different granularities even within the same modality: overly fine-grained entries can dilute context, while overly coarse ones may bundle unrelated information. For example, complex analytical questions may require full documents or videos, while simple factoid questions are better served with a single paragraph or short video clip.

To accommodate this, we further decompose each modality into multiple granularity levels, organizing them into distinct corpora. For example, documents are additionally segmented into paragraphs and stored in a paragraph-level corpus, and similarly, videos are divided into short clips and stored, while images are kept intact since they are inherently piecemeal. Overall, with these modality- and granularity-aware corpora (including paragraphs, documents, tables, images, clips, and videos) in place, as well as an additional no-retrieval option to efficiently handle straightforward queries (that require no external knowledge), our UniversalRAG dynamically routes each query to the most relevant knowledge sources, ultimately supporting the diverse information needs of real-world users.

We validate UniversalRAG on 10 datasets spanning diverse modalities and granularities, where it outperforms all baselines by large margins on average, confirming its effectiveness in handling diverse types of queries. Moreover, UniversalRAG improves efficiency via modality-aware retrieval and appropriate granularity selection, while maintaining robustness on out-of-distribution datasets.

%% file: figures/fig_concept.tex
\vspace{0.35in}
\begin{figure}[h]
    \centering
    \includegraphics[width=\linewidth]{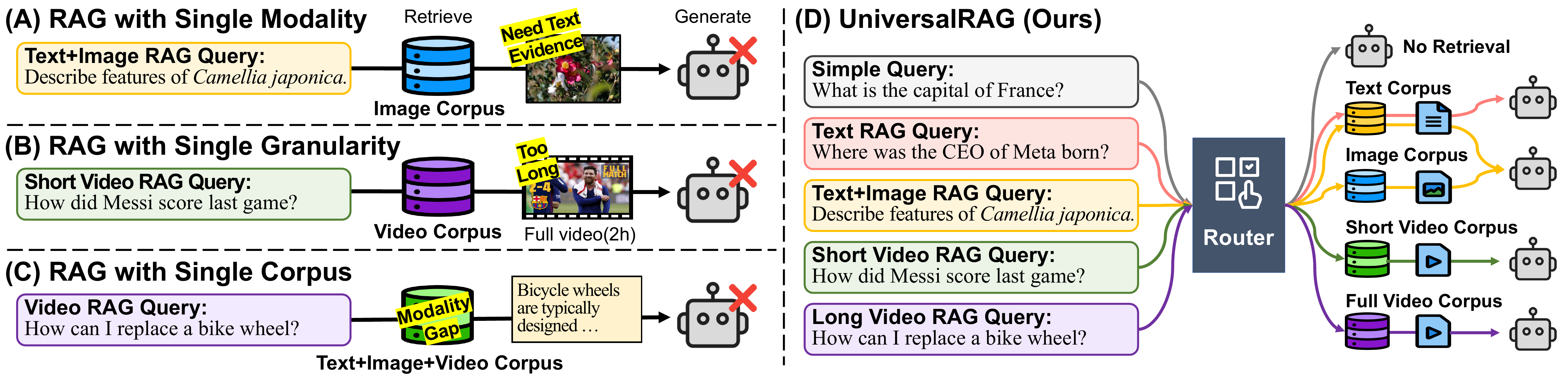}
    \caption{Conceptual illustration comparing existing RAG strategies with our proposed UniversalRAG.}
    \label{fig:fig_concept}
\end{figure}

%% file: figures/fig_embedding.tex
\begin{wrapfigure}{r}{0.42\textwidth}
    \centering
    \vspace{-\intextsep}
    \includegraphics[width=\linewidth]{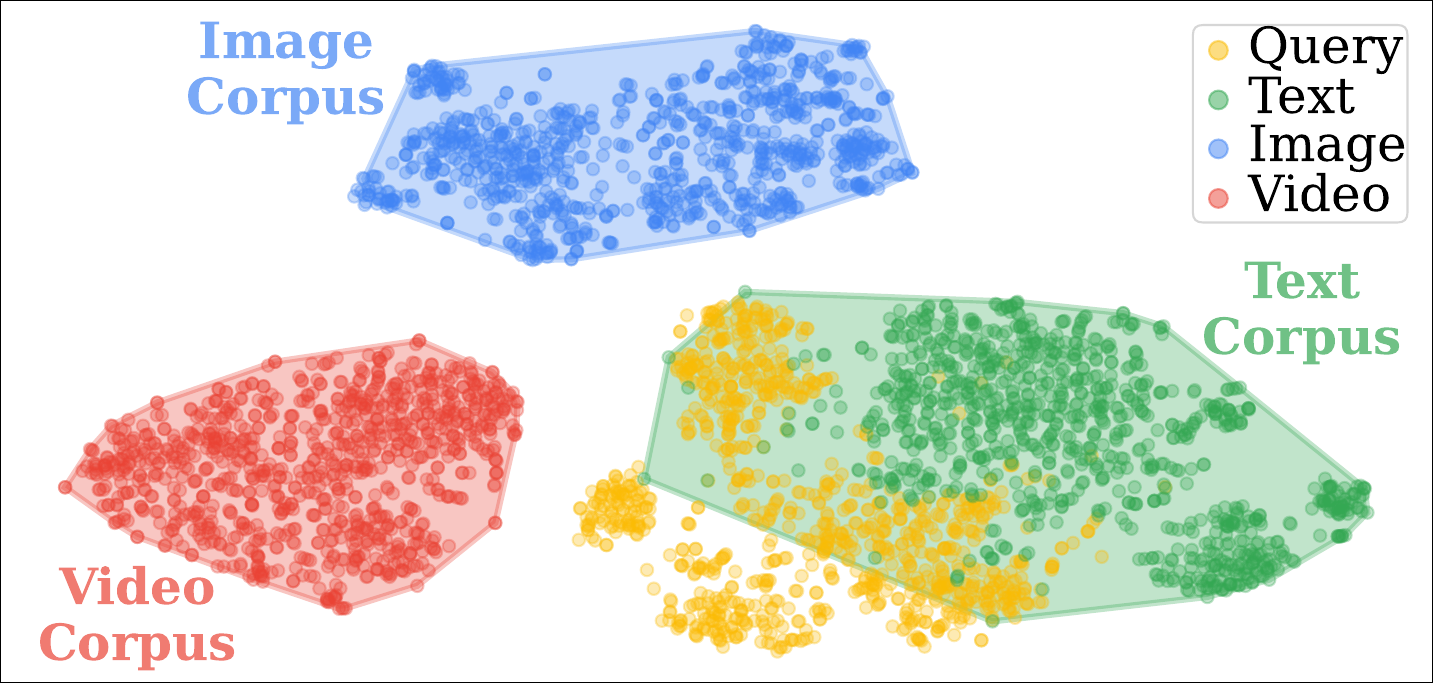}
    \vspace{-0.15in}
    \caption{t-SNE plot of the unified embedding space.}
    \label{fig:embedding}
    \vspace{-0.1in}
\end{wrapfigure}

%% file: sections/2_method.tex
\clearpage
\section{Method}

We begin by describing the preliminaries.

\subsection{Preliminaries}

\paragraph{Large Vision Language Models} 
Let us first define LLMs, which take an input sequence of tokens $\vx = [x_1, x_2, \dots, x_n]$ and generate an output sequence of tokens $\vy = [y_1, y_2, \dots, y_m]$, as follows: $\vy = \texttt{LLM}(\vx)$, where $x_i$ and $y_i$ are represented in text. Building on top of LLMs, Large Vision-Language Models (LVLMs) extend their capability to support multimodal understanding by incorporating visual encoders~\citep{Qwen-VL, InternVL, LLaVA1.5, LLaVA-OneVision}, to process both the textual and visual inputs. Formally, similar to LLMs, LVLMs can be functionalized as $\vy = \texttt{LVLM}(\vx)$, whose input token $x_i$ is extended to either textual or visual. However, although they are extensively trained, LVLMs themselves are limited to their parametric knowledge, and often struggle with queries that require (fine-grained or up-to-date) information, less or not exposed for training.

\paragraph{Retrieval-Augmented Generation}
To address the aforementioned limitations of using only the parametric knowledge, RAG has been widely used, whose core idea is to retrieve query-relevant information from a large corpus and incorporate it into the generation process. Formally, in response to a query $\vq$, a retrieval model $\mathcal{T}$ fetches the relevant context $\vc$ from a corpus $\mathcal{C}$: $\vc = \mathcal{T}(\vq; \mathcal{C})$. Then, in the subsequent generation step, $\texttt{LVLM}$ generates a response $\va$ conditioned on the query and retrieved context: $\va = \texttt{LVLM}(\vq, \vc)$. However, most existing RAG approaches are restricted to retrieving from a single corpus consisting of entries from a single modality (such as only the textual documents), limiting their ability to handle diverse queries with knowledge requirements that vary across them.

\subsection{UniversalRAG}

We introduce UniversalRAG that dynamically identifies and routes queries to the most appropriate modality and granularity for targeted retrieval.

\paragraph{Challenges in Multi-Corpus Retrieval}
To accommodate the diverse knowledge needs of real-world queries, which may involve heterogeneous sources spanning different modalities, we consider a set of modality-specific corpora, where each corpus $\mathcal{C}_m$ contains items of modality $m$. Notably, one straightforward approach to operationalize this is to aggregate all corpora into a unified corpus $\mathcal{C}_{\texttt{unified}} = \bigcup_{m \in M} \mathcal{C}_{m}$ and embed all items into a shared space using a multimodal encoder, as for retrieval over a single corpus: $\vc = \mathcal{T}(\vq; \mathcal{C}_{\texttt{unified}})$. However, we find this approach suffers from modality gap (\cref{fig:embedding,fig:appendix_embedding}), where queries, being textual, align more closely with elements in the text corpus regardless of the modality required. Therefore, instead of forcing all heterogeneous elements into a unified corpus, we propose selectively engaging the most relevant corpora needed for queries.

\paragraph{Modality-Aware Retrieval}
To sidestep the issue of modality gap (introduced by handling all modalities over the unified space), we instead propose to break down the overall retrieval process into two subsequent stages: (1) identifying the most relevant set of modalities for the query; and (2) performing targeted retrieval within the selected modality-specific corpora. Specifically, instead of aggregating all modality-specific corpora, we preserve each corpus in its original form with an independent embedding space. After that, to direct queries to their best-aligned knowledge sources, we introduce a routing module $\mathcal{R}$ that dynamically predicts the modalities best suited for a query $\vq$, yielding $\mathcal{R}(\vq)=M_\vq$ where $M_\vq$ is the set of modalities for $\vq$. Retrieval is then restricted to the corresponding corpora $\{\mathcal{C}_m \mid m \in M_\vq\}$, using any off-the-shelf retriever $\mathcal{T}_m$ tailored to each modality, thereby avoiding the modality gap issue present in a unified space. \cref{prop:unifiedemb} formalizes the advantage of modality-aware routing over unified embeddings, and we provide its proof in \cref{sec:appendix_prop}.

\begin{tcolorbox}
[colback=gray!5, colframe=gray!50, boxrule=0.5pt, arc=2mm, left=4pt, right=4pt, top=3pt, bottom=3pt]
\begin{proposition}
    Let the similarity score in a unified embedding space $\mathcal{C}_{\text{\normalfont\texttt{unified}}}$ be defined as
    \[
        s(\vq, \vc)=\alpha\cdot \mathbf{1}\{m(\vq)=m(\vc)\}+\beta\cdot r(\vq, \vc)+\varepsilon,
    \]
    where $\alpha>0$ is a modality bias, $m(\cdot)$ denotes the modality, and $r(\cdot, \cdot)$ measures semantic relevance. If $\alpha$ is sufficiently large relative to the variance of $r$, the probability of retrieving items from the required modality $m^\ast(\vq)$ is less than under modality-aware routing followed by within-modality retrieval.
    \label{prop:unifiedemb}
\end{proposition}
\end{tcolorbox}
\vspace{0.05in}

\noindent However, while this routing principle mitigates the modality gap, organizing corpora solely by the modality might still be suboptimal since different queries require varying levels of granularity.

\paragraph{Granularity-Aware Retrieval}
To accommodate the varying complexity and information scope of different queries, we extend UniversalRAG to operate not only across modalities but also across different levels of granularity within each modality. To be specific, rather than treating each modality-specific corpus as a flat collection of items, we organize it into representations at multiple resolutions, enabling retrieval to target either fine-grained details or broader context as required by the query. To reflect this richer organization of corpora, the routing module $\mathcal{R}$ expands its prediction space to include modality-granularity pairs best suited to a query, as well as a no-retrieval option for cases where external context is unnecessary:
\begin{equation}
    \mathcal{R}: Q \to \{\varnothing\} \cup \mathcal{P}\left(\bigcup_{m \in M} \{m\}\times G_m\right),
\end{equation}
where $M$ is the set of modalities and $G_m$ is the set of granularities available for modality $m$. Once the router predicts the relevant pairs, retrieval is performed over the corresponding corpora, using retrievers specialized for each modality to obtain the relevant content $\vc$. Finally, the \texttt{LVLM} generates the answer $\va$ with $\vc$, customized to the modality and granularity for each individual query, thereby enabling the universal, one-for-all RAG framework.

\subsection{Router Implementation Strategies}

A key component of UniversalRAG is the router, which is responsible for determining the optimal modality and granularity of knowledge for a query.

\paragraph{Training-based Router}
To perform the routing task, we first consider training the available models to predict the appropriate modality–granularity pair for each query. However, since ground-truth labels (for the modality and granularity the query should be routed to) are not available, we leverage inductive biases in existing benchmarks, mapping each dataset to routing targets that match its task characteristics, allowing us to automatically obtain a labeled corpus without manual annotation. We then train open-source LVLMs to serve as the router using a multi-hot label representation and cross-entropy loss. At inference time, the router produces a sigmoid distribution over modality-granularity pairs and returns all configurations whose scores exceed a predefined threshold, enabling cross-modal and multi-granularity retrieval when necessary.

\paragraph{Training-free Router}
Alternatively, we also explore a training-free approach that leverages the broad knowledge and robust reasoning capabilities of modern frontier models, such as Gemini~\citep{Gemini}. Instead of learning from labeled data, the model is directly prompted to act as a router. To achieve this, we first design the prompt template (used to elicit routing), which describes the objective and includes examples demonstrating how different types of queries correspond to specific retrieval targets (See \cref{fig:prompt_route} for details). Then, at inference, the model is prompted with this template to predict the most suitable modality-granularity pairs from a predefined set. This eliminates the need for supervised labels or task-specific training, offering the flexibility to adapt to new domains.

%% file: sections/3_experiment.tex
\clearpage
\section{Experiment}

\input{tables/tab_main}
\input{figures/fig_metric_comparison}

\subsection{Experimental Setup}
\label{sec:experiment_setup}

We now explain the experimental setup, including datasets, models, and implementation details.

\paragraph{Datasets}
To evaluate UniversalRAG, we compile a comprehensive benchmark covering RAG tasks across seven modalities and granularities. For the no-retrieval setting, we use \textbf{MMLU}~\citep{MMLU}. For text-based RAG, we include \textbf{Natural Questions (NQ)}~\citep{nq} for single-hop, paragraph-level retrieval, and \textbf{HotpotQA}~\citep{hotpotqa} for multi-hop, document-level retrieval. To consider diverse scenarios, we include \textbf{HybridQA}~\citep{hybridqa} for reasoning over text and tables, \textbf{MRAG-Bench (MRAG)}~\citep{mrag-bench} for image RAG, and \textbf{WebQA}~\citep{WebQA} and \textbf{InfoSeek}~\citep{infoseek} for cross-modal RAG over text and images. Lastly, for RAG with videos, \textbf{LVBench}~\citep{lvbench} is used for queries over short or localized video segments, as well as \textbf{VideoRAG-Wiki} and \textbf{VideoRAG-Synth}~\citep{VideoRAG} for queries grounded on long-form or complete videos. Please refer to \cref{sec:appendix_dataset} for more details.

\vspace{-0.04in}

\paragraph{Knowledge Corpora}
To support the aforementioned, diverse RAG scenarios with various modalities and granularities, we consider their corresponding corpora. Recall that we define seven routing pathways: \textbf{None}, \textbf{Paragraph}, \textbf{Document}, \textbf{Table}, \textbf{Image}, \textbf{Clip}, and \textbf{Video}, with cross-modal routing allowing queries to span multiple modalities. For the paragraph and document corpora, we use Wikipedia at the levels of paragraphs~\citep{Wikipedia} and documents~\citep{longrag}. The table corpus is built by collecting tables from the HybridQA benchmark. For the image, we adopt corpora from MRAG-Bench, WebQA, and InfoSeek datasets. Lastly, we construct two video corpora at different scales: a video-level corpus consisting of full-length videos from LVBench and VideoRAG datasets, and a clip-level constructed by segmenting these videos into multiple short clips.

\paragraph{Methods}
\label{sec:models}
We compare our UniversalRAG to a diverse set of 12 baselines, grouped into four categories. The first is \textbf{Naïve}, which directly answers queries without retrieving external knowledge. In addition, the group of \textbf{Unimodal RAGs} includes \textbf{ParagraphRAG}, \textbf{DocumentRAG}, \textbf{TableRAG}, \textbf{ImageRAG}, \textbf{ClipRAG}, and \textbf{VideoRAG} methods, which retrieve information exclusively from their respective corpora and leverage it for response generation. The third group of \textbf{Unified Embedding Multimodal RAGs} uses multimodal encoders to align different modalities into a shared embedding space for retrieval, including \textbf{UniRAG}~\citep{unirag}, \textbf{GME}~\citep{GME}, $\textbf{PE}_{\textbf{core}}$~\citep{pecore}, and \textbf{VLM2Vec-V2}~\citep{vlm2vecv2}. \textbf{MultiRAG} is included in the last group of \textbf{Multi-corpus Multimodal RAGs}, which performs retrieval over all the available corpora and incorporates the retrieved results for response generation. Notably, as \textbf{UniversalRAG} can be operationalized with different routing strategies, we implement \textbf{training-based variants}, which leverage Qwen3-VL-2B-Instruct~\citep{qwen3vl}, InternVL3.5-1B~\citep{internvl3_5}, and T5Gemma 2 270M~\citep{t5gemma} (finetuned on the automatically constructed routing dataset), as well as \textbf{training-free variants}, which prompt GPT-5~\citep{gpt5} and Qwen3-VL-8B-Instruct~\citep{qwen3vl} to select appropriate modality-granularity pairs. Finally, we include an oracle setup (\textbf{Oracle}), which routes each query to its ideal corpora, non-comparable with others.

\paragraph{Implementation Details}
For response generation, we utilize multiple LVLMs, Qwen3-VL-8B-Instruct~\citep{qwen3vl}, InternVL3.5-8B~\citep{internvl3_5}, and Molmo2-4B~\citep{molmo2}. Also, to take advantage of UniversalRAG in routing the retrieval process to the modality-specific corpus, we use modality-specific encoders: Qwen3-Embedding-4B~\citep{qwen3embedding} for text, VLM2Vec-V2~\citep{vlm2vecv2} for vision, and dense row-level embedding~\citep{targettable} with the text encoder for tables. We provide further details (including router training) in \cref{sec:appendix_implementation}.

\input{figures/fig_ablation}
\input{tables/tab_ablation1}

\subsection{Experimental Results and Analyses}
\label{sec:experimental_results}

Now we present the overall results across diverse RAG scenarios, followed by a detailed analysis.

\paragraph{Overall Results}
We present the modality- and granularity-specific results in \cref{tab:main}, along with the averaged results with different LVLMs in \cref{fig:average_scores}, from which we observe that UniversalRAG consistently achieves the best performance on average. Specifically, in \cref{tab:main}, the results compared against the unimodal RAG baselines corroborate our hypothesis that retrieving from the modality (or granularity) that aligns best with the information needs of the queries achieves the highest accuracy; however, mismatches between the query and retrieval source results in significant degradation, which supports our claim that considering diverse modalities in the universal workflow is necessary for realistic RAG. Also, the level of granularity within each modality affects performance, suggesting that fine-grained retrieval and generation are necessary. In addition to them, UniversalRAG significantly outperforms unified embedding multimodal RAG baselines, confirming the issue of the modality gap inherent within them (See \cref{fig:embedding,fig:appendix_embedding}). Lastly, when compared with the MultiRAG baseline (within the multi-corpus multimodal RAG category), which results in suboptimal performance due to the inclusion of noise from irrelevant modalities in generation, our UniversalRAG remains effective. Its strong performance is due to its core idea around modality-aware routing, enabling the dynamic retrieval from the most relevant modalities and granularities for each query, yielding performance gains despite using several corpora.

\paragraph{Effectiveness of Cross-Modal Retrieval}
While many queries can be addressed by using a single, most prominent modality, certain tasks benefit from integrating evidence across multiple modalities. For instance, HybridQA requires reasoning that spans both structured tables and accompanying textual sources, while WebQA involves visually grounded questions that pair text with images. \cref{tab:cross_retrieval} shows that, compared to uni-modal retrieval, for which each query is routed to a single relevant source, cross-modal retrieval achieves consistently stronger performance. By enabling queries to be routed across multiple modalities, the cross-modal retrieval can leverage complementary evidence that would otherwise be missed by the uni-modal approach. These highlight the effectiveness of UniversalRAG’s flexible routing mechanism, which dynamically retrieves information from multiple sources rather than relying on a single modality.

\paragraph{Effectiveness of Modality Routing}
To investigate the effectiveness of our routing method, we compare the distribution of retrieved modalities for VLM2Vec-V2, GME, and UniversalRAG (with Qwen3-VL-2B) in \cref{fig:modality_selection_rate}. Using 200 sampled queries per benchmark and normalizing distributions, we find that VLM2Vec-V2 retrieves exclusively text, while GME similarly exhibits a strong bias toward text regardless of the query’s required modality, reflecting the modality gap inherent to unified embedding spaces. In contrast, UniversalRAG retrieves more evenly across modalities, indicating that the router effectively mitigates modality bias and adaptively selects appropriate knowledge sources. This leads to higher modality retrieval accuracy, and consequently, higher retrieval recall, as shown in \cref{tab:recall}. While GME achieves comparable recall on text and image corpora, its inability to accurately retrieve from the correct modality leads to lower recall on multimodal corpora that include videos. Yet, UniversalRAG consistently retrieves from the correct modality, enabling it to achieve higher recall than baselines across all scenarios.

\input{tables/tab_ablation2}

\paragraph{Effectiveness of Multigranularity}
Given the observed benefits of corpus selection in \cref{tab:main}, we investigate its impact beyond modality by comparing UniversalRAG at varying levels of granularity\footnote{In our main experiments, we adopt a binary level of granularity to strike a balance between effectiveness and efficiency.}. \cref{tab:granularity_perf} shows that incorporating granularity-aware corpus selection leads to consistent performance gains by avoiding the retrieval of context that is either insufficient (e.g., a short paragraph lacking key entities for multi-hop reasoning) or excessive (e.g., a full video when only a short clip is relevant), both of which can hinder accurate response generation. Also, as additional granularity levels are introduced, we observe further improvements in some cases, though gains are not strictly monotonic across tasks, reflecting the trade-off between context sufficiency and noise. Please see \cref{sec:appendix_theory_granularity} for a theoretical analysis supporting these findings.

\paragraph{Efficiency of Modality-Specific Retrieval}
Beyond accuracy, UniversalRAG also improves efficiency by reducing the search space: it leverages modality- and granularity-aware routing to restrict retrieval to only the most relevant sources, instead of querying a unified embedding index that aggregates all modalities into a single mega-corpus. Also, the overhead for routing is small as this cost is outweighed at scale by the size of the search space, leading to sub-linear latency growth as corpus size increases, as shown in \cref{fig:retrieval_latency}. Here, UniversalRAG eventually achieves lower latency than unified embedding methods at large corpus sizes, with the gap widening further at very large scales (beyond 10M entries). This scalability makes UniversalRAG a practical solution for real-world applications, where corpora are significantly larger than our experimental settings. We provide an in-depth theoretical analysis of efficiency in \cref{sec:appendix_theory_efficiency}.

\paragraph{Analysis on Router Size}
To examine whether the routing cost can be further reduced by using smaller models as routers without sacrificing accuracy, we train three models~\citep{internvl3_5,t5gemma,smolvlm} ranging from 256M to 4B parameters and measure router accuracy. As shown in \cref{fig:router_model_size}, router accuracy consistently improves with increasing model size within each architecture, suggesting the scalability of our routing approach. While the largest models achieve near-perfect routing performance, a 1B-parameter model attains approximately 90\% accuracy, indicating that compact models can serve as effective routers in UniversalRAG.

\input{tables/tab_casestudy}

\paragraph{Generalizability on Out-of-Domain Scenarios}
As shown in \cref{tab:main}, UniversalRAG with trained routers outperforms the training-free router (sometimes even approaching oracle performance), and a natural follow-up question is how these routers behave on unseen, out-of-domain (OOD) datasets. To investigate this, we evaluate on six OOD datasets (detailed in \cref{sec:appendix_dataset_ood}), with results presented in \cref{tab:route_generation,tab:appendix_main_ood}. In contrast to the in-domain setting, trained routers exhibit noticeable performance degradation, whereas the training-free router generalizes robustly and even surpasses the trained variants. Nevertheless, UniversalRAG remains effective in OOD scenarios and consistently outperforms all baselines, including those using the unified embedding spaces or random modality and granularity assignment, highlighting the benefit of adaptive, modality- and granularity-aware retrieval.

\paragraph{Ensemble Strategy for Robust Routing}
Building on the trade-off between the high in-domain accuracy of trained routers and the strong OOD generalization of training-free routers, we propose ensemble strategies that leverage their complementary strengths. Specifically, we explore confidence-based ensembling, which uses the trained router’s prediction when its confidence exceeds a threshold and otherwise falls back to the training-free router, as well as majority voting, which selects the majority prediction from three routers (training-based and free) with random tie-breaking. \cref{tab:route_generation} shows that UniversalRAG with the ensemble routing achieves a robust balance between accuracy and generalization, making it well suited for real-world scenarios with unseen or shifting distributions.

\paragraph{Case Study}
We present a case study of UniversalRAG in \cref{tab:casestudy}. The query asks for the number of statues of people on the Michigan Soldiers and Sailors Monument. Both TextRAG and ImageRAG retrieve the relevant and correct evidence; however, each modality alone is insufficient to determine the full count. TextRAG lacks the information needed to aggregate all statues, while ImageRAG suffers from partial occlusion. VideoRAG fails to retrieve relevant evidence, as the video corpus does not contain information useful for this query. In contrast, UniversalRAG routes the query to both the ``Paragraph'' and ``Image'' corpora, allowing cross-modal reasoning and correctly identifying all nine statues. More case studies are provided in \cref{sec:qualitative_results}.

%% file: tables/tab_main.tex
\begin{table*}[t]
    \centering
    \caption{Results of diverse RAG methods with Qwen3-VL-8B-Instruct across modalities. \textbf{Bold} denotes the best performance and \underline{underlined} indicates the second-best among UniversalRAG variants, using either \textcolor{teal!80}{trained} or \textcolor{blue!80}{training-free} routers. R-L and BERT correspond to ROUGE-L and BERTScore, respectively.}
    \vspace{-0.05in}
    \renewcommand{\arraystretch}{1.0}
    \renewcommand{\tabcolsep}{1.6mm}
    \small
    \resizebox{\textwidth}{!}{
        \begin{tabular}{l c cc cc cc c cc c c cc cc c}
            \toprule
            \multicolumn{1}{c}{} & \textbf{MMLU} & \multicolumn{2}{c}{\textbf{NQ}} & \multicolumn{2}{c}{\textbf{HotpotQA}} & \multicolumn{2}{c}{\textbf{HybridQA}} & \textbf{MRAG} & \multicolumn{2}{c}{\textbf{WebQA}} & \textbf{InfoSeek} & \textbf{LVBench} & \multicolumn{2}{c}{\textbf{VideoRAG-Wiki}} & \multicolumn{2}{c}{\textbf{VideoRAG-Synth}} & \multirow[c]{2}{*}[-0.3em]{\textbf{Avg}} \\
            \cmidrule(lr){2-2} \cmidrule(lr){3-4} \cmidrule(lr){5-6} \cmidrule(lr){7-8} \cmidrule(lr){9-9} \cmidrule(lr){10-11} \cmidrule(lr){12-12} \cmidrule(lr){13-13} \cmidrule(lr){14-15} \cmidrule(lr){16-17}
            \textbf{Models} & Acc & EM & F1 & EM & F1 & EM & F1 & Acc & R-L & BERT & Acc & Acc & R-L & BERT & R-L & BERT \\
            \midrule
            \midrule
            Naïve & 74.39 & 18.85 & 28.98 & 21.10 & 29.53 & 2.80 & 7.81 & 49.22 & 58.12 & 93.78 & 18.10 & 28.83 & 19.78 & 86.51 & 35.86 & 90.76 & 35.59 \\
            \noalign{\vskip 0.25ex}\cdashline{1-18}\noalign{\vskip 0.75ex}
            ParagraphRAG & 74.39 & 39.25 & 51.32 & 23.40 & 31.45 & 5.10 & 9.21 & 46.71 & 51.70 & 92.53 & 19.75 & 24.07 & 17.62 & 85.94 & 32.86 & 89.97 & 37.26 \\
            DocumentRAG & 71.29 & 21.95 & 30.26 & 26.35 & 34.72 & 3.75 & 7.40 & 43.68 & 45.57 & 91.50 & 16.80 & 15.70 & 16.54 & 85.60 & 31.18 & 89.61 & 32.26 \\
            TableRAG & 72.51 & 11.80 & 18.73 & 16.45 & 22.28 & 9.65 & 13.86 & 43.39 & 44.75 & 91.47 & 9.15 & 16.47 & 12.16 & 84.04 & 30.67 & 89.45 & 29.45 \\
            ImageRAG & 73.33 & 17.15 & 25.37 & 19.15 & 26.15 & 2.20 & 5.69 & 52.55 & 67.96 & 95.65 & 20.15 & 25.35 & 19.50 & 87.06 & 36.10 & 90.77 & 35.04 \\
            ClipRAG & 73.33 & 16.70 & 24.56 & 19.30 & 26.75 & 2.35 & 6.16 & 48.93 & 65.68 & 94.83 & 9.85 & 33.72 & 21.10 & 87.67 & 39.39 & 91.47 & 35.18 \\
            VideoRAG & 74.91 & 15.85 & 23.69 & 20.00 & 27.02 & 2.30 & 5.78 & 48.04 & 64.97 & 94.67 & 11.25 & 32.05 & 20.89 & 87.65 & 40.05 & 91.54 & 35.01 \\
            \noalign{\vskip 0.25ex}\cdashline{1-18}\noalign{\vskip 0.75ex}
            UniRAG & 70.06 & 19.30 & 29.71 & 19.35 & 26.89 & 2.85 & 7.89 & 44.86 & 53.26 & 92.89 & 19.05 & 22.65 & 18.05 & 86.11 & 32.41 & 89.68 & 32.93 \\
            GME & 70.41 & 20.05 & 29.91 & 19.50 & 26.93 & 3.00 & 8.00 & 49.45 & 55.03 & 93.26 & 19.20 & 23.68 & 18.01 & 86.03 & 33.02 & 89.95 & 33.88 \\
            $\text{PE}_{\text{core}}$ & 72.11 & 19.65 & 29.77 & 19.00 & 26.32 & 3.05 & 8.02 & 49.15 & 54.79 & 93.07 & 19.10 & 23.04 & 18.24 & 86.64 & 32.75 & 89.80 & 33.86 \\
            VLM2Vec-V2 & 71.70 & 19.95 & 29.88 & 18.50 & 25.24 & 2.95 & 8.04 & 46.78 & 52.35 & 92.60 & 18.80 & 23.55 & 18.03 & 86.07 & 33.38 & 90.19 & 33.31 \\
            \noalign{\vskip 0.25ex}\cdashline{1-18}\noalign{\vskip 0.75ex}
            MultiRAG & 70.82 & 20.90 & 30.02 & 22.65 & 30.74 & 4.35 & 8.47 & 45.01 & 56.73 & 93.31 & 19.05 & 23.55 & 17.89 & 85.91 & 34.24 & 90.33 & 34.07 \\
            \noalign{\vskip 0.25ex}\cdashline{1-18}\noalign{\vskip 0.75ex}
            \multicolumn{18}{l}{\textbf{UniversalRAG (Ours)}} \\
            \addlinespace[0.2ex]
            \rowcolor{green!8}\multicolumn{18}{l}{\hspace{1em}\textbf{\textit{Trained Routers}}} \\
            \rowcolor{green!8}\hspace{2.5em}Qwen3-VL-2B-Instruct & \underline{74.39} & \underline{38.65} & \underline{50.61} & \textbf{26.10} & \textbf{34.61} & \textbf{11.05} & \textbf{16.23} & \textbf{52.55} & \textbf{70.22} & \textbf{95.86} & \underline{23.20} & \textbf{33.72} & \textbf{20.86} & \textbf{87.63} & \textbf{39.95} & \underline{91.51} & \textbf{42.40} \\
            \rowcolor{green!8}\hspace{2.5em}InternVL3.5-1B & \underline{74.39} & \textbf{38.70} & 50.60 & 25.85 & \underline{34.29} & \underline{10.25} & \underline{14.79} & \textbf{52.55} & \underline{69.14} & \underline{95.72} & \textbf{23.35} & \textbf{33.72} & \underline{20.85} & \textbf{87.63} & \underline{39.90} & \textbf{91.52} & \underline{42.12} \\
            \rowcolor{green!8}\hspace{2.5em}T5Gemma 2 270M & \textbf{74.62} & \underline{38.65} & \textbf{50.62} & \underline{25.90} & 33.94 & 9.95 & 14.70 & 50.33 & 69.03 & 95.66 & 21.95 & 33.59 & 20.81 & 87.61 & 39.43 & 91.38 & 41.68 \\            
            \addlinespace[0.2ex]
            \rowcolor{blue!8}\multicolumn{18}{l}{\hspace{1em}\textbf{\textit{Training-free Routers}}} \\
            \rowcolor{blue!8}\hspace{2.5em}GPT-5 & 74.27 & 34.50 & 46.21 & 24.35 & 32.71 & 4.95 & 8.79 & 50.11 & 62.38 & 94.52 & 21.45 & 32.30 & 19.61 & 86.42 & 35.94 & 90.69 & 39.26 \\
            \rowcolor{blue!8}\hspace{2.5em}Qwen3-VL-8B-Instruct & 74.09 & 35.20 & 47.09 & 24.65 & 33.12 & 5.25 & 9.44 & 50.04 & 65.27 & 94.77 & 20.65 & 32.43 & 18.24 & 86.07 & 34.77 & 90.11 & 39.46 \\
            \noalign{\vskip 0.25ex}\cdashline{1-18}\noalign{\vskip 0.75ex}
            Oracle & 74.39 & 39.25 & 51.32 & 26.35 & 34.72 & 10.55 & 15.20 & 52.55 & 71.17 & 96.02 & 23.35 & 33.72 & 20.89 & 87.65 & 40.05 & 91.54 & 42.45 \\
            \bottomrule
        \end{tabular}
    }
    \label{tab:main}
\end{table*}

%% file: figures/fig_metric_comparison.tex
\begin{figure*}[t!]
    \centering
    \includegraphics[width=0.97\textwidth]{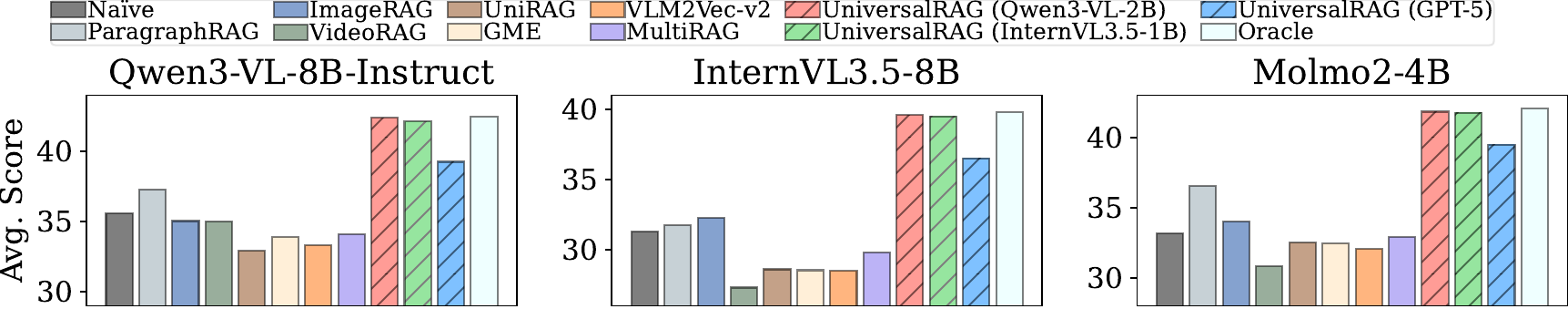}
    \vspace{-0.05in}
    \caption{Comparison of averaged evaluation results across different RAG methods and LVLMs.}
    \label{fig:average_scores}
    \vspace{-0.05in}
\end{figure*}

%% file: figures/fig_ablation.tex
\begin{figure*}[t!]
    \centering
    \begin{minipage}[t]{0.32\linewidth}
        \centering
        \includegraphics[width=0.975\columnwidth, trim=0 0 0 1, clip]{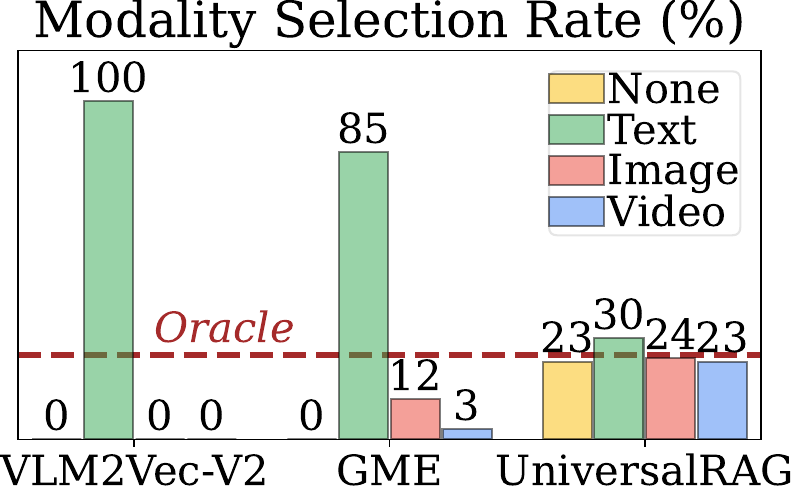}
        \vspace{-0.1in}
        \captionof{figure}{Distribution of the retrieved data modalities.}
        \label{fig:modality_selection_rate}
    \end{minipage}
    \hfill
    \begin{minipage}[t]{0.32\linewidth}
        \centering
        \includegraphics[width=0.95\columnwidth]{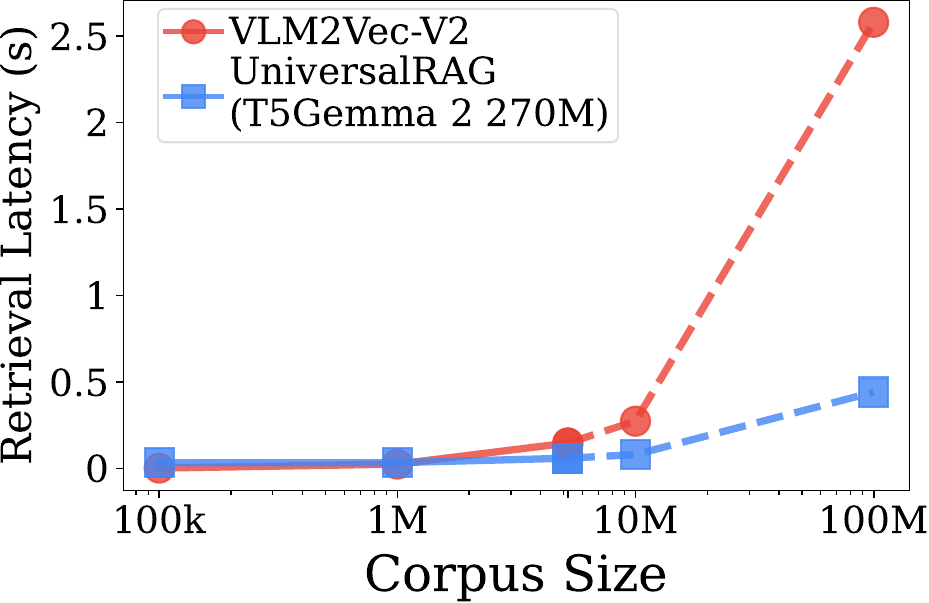}
        \vspace{-0.1in}
        \captionof{figure}{Retrieval latency per query across corpus sizes.}
        \label{fig:retrieval_latency}
    \end{minipage}
    \hfill
    \begin{minipage}[t]{0.32\linewidth}
        \centering
        \includegraphics[width=0.975\columnwidth]{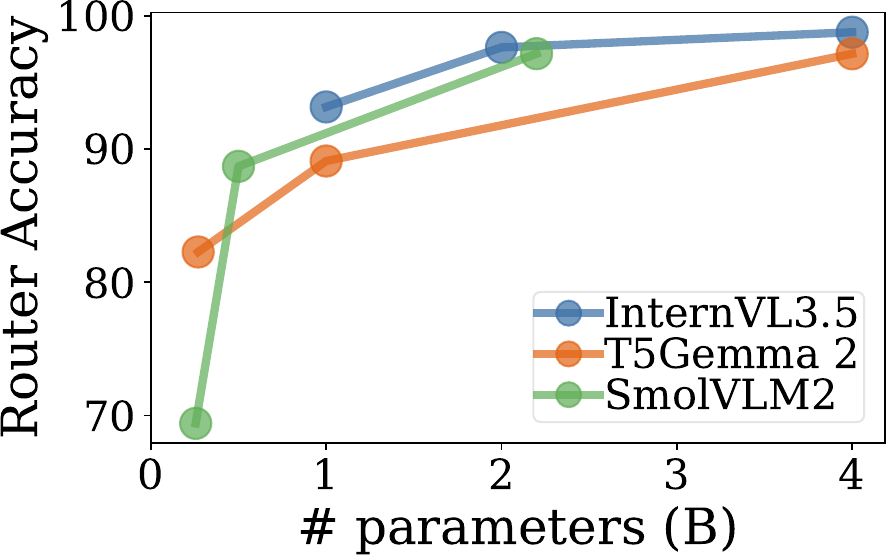}
        \vspace{-0.1in}
        \captionof{figure}{Router accuracy with varying the router model size.}
        \label{fig:router_model_size}
    \end{minipage}
\end{figure*}

%% file: tables/tab_ablation1.tex
\begin{figure*}[t!]
    \centering
    \begin{minipage}[t]{0.49\linewidth}
        \centering
        \captionof{table}{Performance comparison of uni-modal and cross-modal approaches across different router models. Among models, GPT-5 is the only training-free router.}
        \vspace{-0.1in}
        \renewcommand{\arraystretch}{1.09}
        \renewcommand{\tabcolsep}{2.0mm}
        \small
        \resizebox{\linewidth}{!}{
            \begin{tabular}{l l c c c c}
                \toprule
                & & \multicolumn{2}{c}{\textbf{HybridQA}} & \multicolumn{2}{c}{\textbf{WebQA}} \\
                \cmidrule(lr){3-4} \cmidrule(lr){5-6}
                \textbf{Models} & \textbf{Retrieval} & EM & F1 & R-L & BERT \\
                \midrule
                \midrule
                \multirow{2}{*}{Qwen3-VL-2B}
                 & Uni-modal & 9.60 & 14.56 & 67.93 & 95.58 \\
                 & Cross-modal & \textbf{11.05} & \textbf{16.23} & \textbf{70.22} & \textbf{95.86} \\
                \noalign{\vskip 0.25ex}\cdashline{1-6}\noalign{\vskip 0.75ex}
                \multirow{2}{*}{InternVL3.5-1B}
                 & Uni-modal & 9.65 & 13.86 & 67.90 & 95.49 \\
                 & Cross-modal & \textbf{10.25} & \textbf{14.79} & \textbf{69.14} & \textbf{95.72} \\
                \noalign{\vskip 0.25ex}\cdashline{1-6}\noalign{\vskip 0.75ex}
                \multirow{2}{*}{GPT-5}
                 & Uni-modal & 4.75 & 8.57 & 60.54 & 94.04 \\
                 & Cross-modal & \textbf{4.95} & \textbf{8.79} & \textbf{62.38} & \textbf{94.52} \\
                \bottomrule
            \end{tabular}
        }
        \label{tab:cross_retrieval}
    \end{minipage}
    \hfill
    \begin{minipage}[t]{0.49\linewidth}
        \centering
        \captionof{table}{Modality accuracy (in corpus selection) and recall of retrieved items for retrieval methods. Among UniversalRAG variants, GPT-5 is only training-free router.}
        \vspace{-0.1in}
        \renewcommand{\arraystretch}{0.92}
        \renewcommand{\tabcolsep}{1mm}
        \small
        \resizebox{\linewidth}{!}{
            \begin{tabular}{l c ccc}
                    \toprule
                    & \multirow{2}{*}{\raisebox{-2.1ex}{\makecell{\textbf{Modality}\\ \textbf{Acc}}}} & \multicolumn{3}{c}{\textbf{Recall}} \\ 
                    \cmidrule(lr){3-5}
                    \textbf{Models} & & R@1 & R@3 & R@5 \\
                    \midrule
                    \midrule
                    UniRAG & 25.00 & 0.01 & 0.03 & 0.04 \\
                    GME & 36.27 & 13.84 & 17.79 & 22.16 \\
                    $\text{PE}_{\text{core}}$ & 25.00 & 0.67 & 1.20 & 1.85 \\
                    VLM2Vec-V2 & 25.00 & 2.30 & 3.69 & 4.12 \\
                    \noalign{\vskip 0.25ex}\cdashline{1-5}\noalign{\vskip 0.75ex}
                    \textbf{UniversalRAG \scriptsize{(Qwen3-VL-2B)}} & \textbf{95.28} & \textbf{21.38} & \textbf{36.29} & \textbf{44.82} \\
                    \textbf{UniversalRAG \scriptsize{(InternVL3.5-1B)}} & \underline{92.39} & \underline{19.66} & \underline{31.82} & \underline{39.20} \\
                    \textbf{UniversalRAG \scriptsize{(GPT-5)}} & 68.22 & 16.33 & 23.72 & 31.41 \\
                    \bottomrule
                \end{tabular}
        }
        \label{tab:recall}
    \end{minipage}
    \vspace{-0.1in}
\end{figure*}

%% file: tables/tab_ablation2.tex
\begin{figure*}[t!]
    \centering
        \begin{minipage}[t]{0.49\linewidth}
        \centering
        \captionof{table}{Performance across different numbers of granularity (\#Gn) for training-free router models. The prompt used to route to finer granularities is shown in \cref{fig:prompt_route_gran}.}
        \vspace{-0.1in}
        \renewcommand{\arraystretch}{0.93}
        \renewcommand{\tabcolsep}{3mm}
        \small
        \resizebox{\linewidth}{!}{
            \begin{tabular}{l c c c c}
                \toprule
                & & \multicolumn{2}{c}{\textbf{HotpotQA}} & \textbf{LVBench} \\
                \cmidrule(lr){3-4} \cmidrule(lr){5-5}
                \textbf{Models} & \textbf{\#Gn} & EM & F1 & Acc \\
                \midrule
                \midrule
                \multirow{4}{*}{GPT-5}
                 & 1 & 23.20 & 31.38 & 31.92 \\
                 & 2 & \underline{24.35} & \underline{32.71} & 32.30 \\
                 & 3 & 24.20 & 32.64 & \underline{32.43} \\
                 & 4 & \textbf{24.70} & \textbf{33.25} & \textbf{32.85} \\
                \noalign{\vskip 0.25ex}\cdashline{1-5}\noalign{\vskip 0.75ex}
                \multirow{4}{*}{Qwen3-VL-8B}
                 & 1 & 23.85 & 32.54 & 31.53 \\
                 & 2 & 24.65 & 33.12 & 32.43 \\
                 & 3 & \underline{24.70} & \underline{33.23} & \underline{32.82} \\
                 & 4 & \textbf{25.05} & \textbf{33.70} & \textbf{33.20} \\
                \bottomrule
            \end{tabular}
        }
        \label{tab:granularity_perf}
        \end{minipage}
    \hfill
    \begin{minipage}[t]{0.49\linewidth}
        \centering
        \captionof{table}{Router accuracy and generation performance across retrieval methods on two settings. Among UniversalRAG variants, GPT-5 is the only training-free router.}
        \vspace{-0.1in}
        \renewcommand{\arraystretch}{0.9}
        \renewcommand{\tabcolsep}{2mm}
        \small
        \resizebox{\linewidth}{!}{
            \begin{tabular}{l c c cc}
                    \toprule
                    & \multicolumn{2}{c}{\textbf{In-Domain}} & \multicolumn{2}{c}{\textbf{Out-Domain}} \\ 
                    \cmidrule(lr){2-3} \cmidrule(lr){4-5}
                    & \multirow{2}{*}{\raisebox{-1.4ex}{\makecell{Router\\ Acc}}} & \multirow{2}{*}{\raisebox{-1.4ex}{\makecell{Avg\\ Score}}} & \multirow{2}{*}{\raisebox{-1.4ex}{\makecell{Router\\ Acc}}} & \multirow{2}{*}{\raisebox{-1.4ex}{\makecell{Avg\\ Score}}} \\
                    \textbf{Models} & & & & \\
                    \midrule
                    \midrule
                    Random & 14.29 & 31.75 & 14.29 & 37.85 \\
                    \noalign{\vskip 0.25ex}\cdashline{1-5}\noalign{\vskip 0.75ex}
                    $\text{PE}_{\text{core}}$ & - & 33.86 & - & 39.08 \\
                    VLM2Vec-V2 & - & 33.31 & - & 38.99 \\
                    \noalign{\vskip 0.25ex}\cdashline{1-5}\noalign{\vskip 0.75ex}
                    \textbf{UniversalRAG \scriptsize{(Qwen3-VL-2B)}} & 95.81 & 42.40 & 71.29 & 44.07 \\
                    \textbf{UniversalRAG \scriptsize{(InternVL3.5-1B)}} & 93.16 & 42.12 & 67.85 & 43.80 \\
                    \textbf{UniversalRAG \scriptsize{(GPT-5)}} & 72.33 & 41.68 & 77.38 & 44.39 \\
                    \noalign{\vskip 0.25ex}\cdashline{1-5}\noalign{\vskip 0.75ex}
                    Ensemble \scriptsize{(Confidence-based)} & \underline{96.02} & \underline{42.53} & \textbf{80.71} & \textbf{44.71} \\
                    Ensemble \scriptsize{(Majority Voting)} & \textbf{98.33} & \textbf{42.83} & \underline{78.56} & \underline{44.54} \\
                    \bottomrule
                \end{tabular}
        }
        \label{tab:route_generation}
    \end{minipage}
    \vspace{-0.075in}
\end{figure*}

%% file: tables/tab_casestudy.tex
\begin{table*}[t]
    \centering
    \renewcommand{\arraystretch}{1.0}
    \renewcommand{\tabcolsep}{1.0mm}
    \caption{Case study comparing unimodal RAGs with fixed modality and granularity against UniversalRAG (Ours).}
    \vspace{-0.1in}
    \scriptsize
    \begin{tabular}{p{0.085\linewidth} p{0.38\linewidth} p{0.085\linewidth} p{0.38\linewidth}}
        \toprule
        \cellcolor{blue!5}\textbf{Question} & \multicolumn{3}{l}{\cellcolor{blue!5}\makecell[l]{How many statues of people are there on the Michigan Soldiers Sailors monument?~~~~~~~~~~~~~~~~~~~~~~~~~\textbf{Answer:} Nine statues of people.}~~~~~~~~~~~~~~~~~~~~~~}\\
        \midrule
        \multirow[t]{2}{*}{\textbf{TextRAG}} & \textbf{Retrieved:} the next section which is surmounted by four male figures depicting the Navy, Infantry, Cavalry, and Artillery branches of the United States Army. Four female allegorical figures, resting on pedestals, are above the male statues and ... & \multirow[t]{2}{*}{\textbf{ImageRAG}} & \textbf{Retrieved: } \hspace{0.05in} \raisebox{-28pt}[0pt][0pt]{\includegraphics[width=0.055\textwidth]{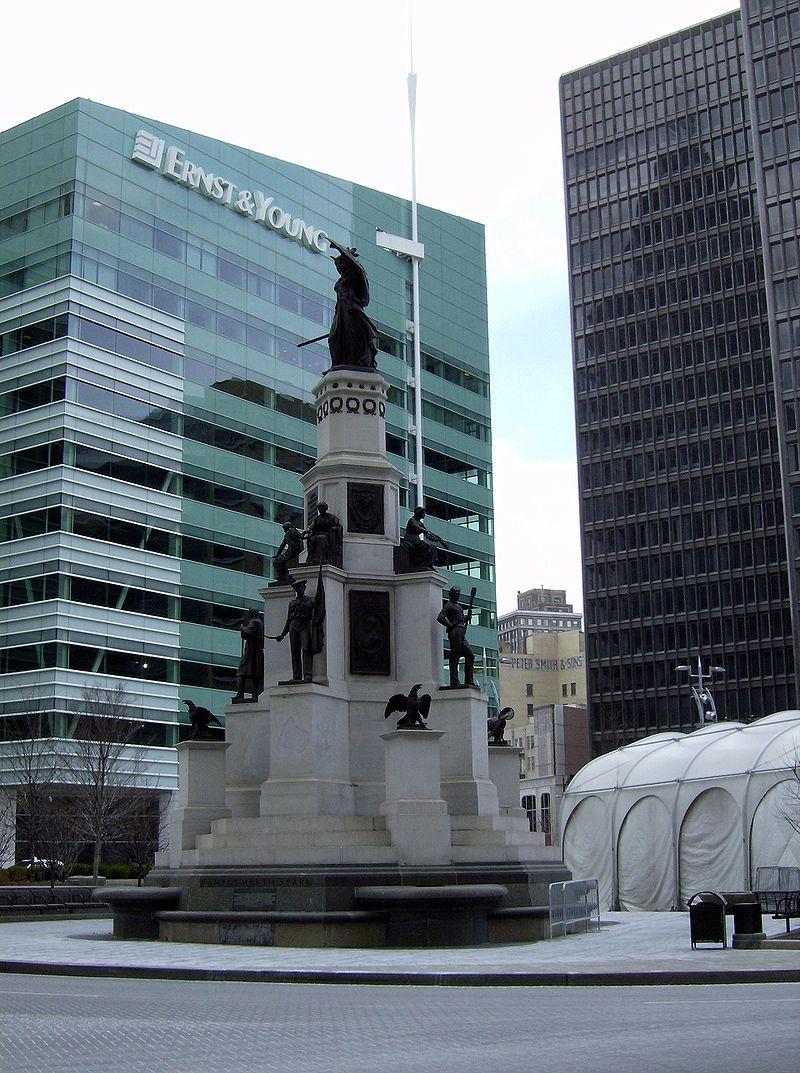}} \\
        & \vspace{0.5pt} \textbf{Response:} Eight people \redx & & \vspace{1pt} \textbf{Response:} Six people \redx\\
        \midrule
        \multirow[t]{2}{*}{\textbf{VideoRAG}} & \textbf{Retrieved:} & \multirow[t]{2}{*}{\textbf{Ours}} & \textbf{Routed to}: Paragraph+Image \\
        & \filmboxsmall{'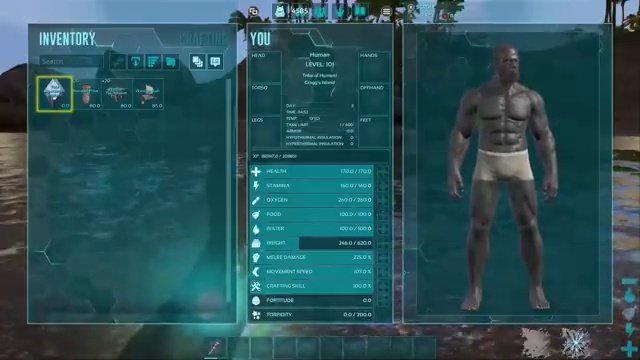'}\filmboxsmall{'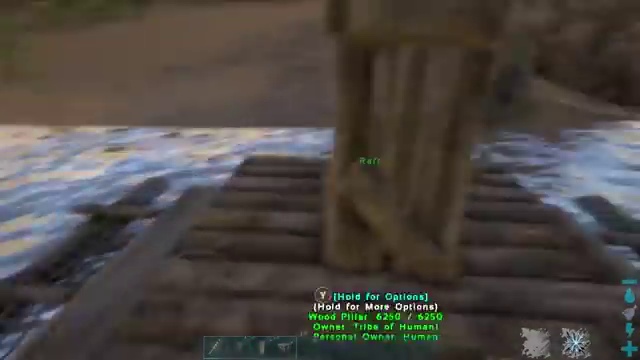'}\filmboxsmall{'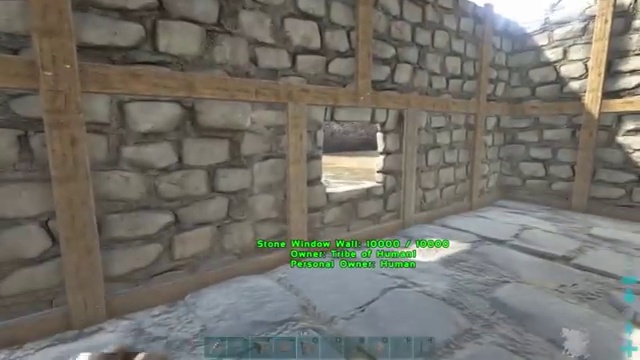'}\filmboxsmall{'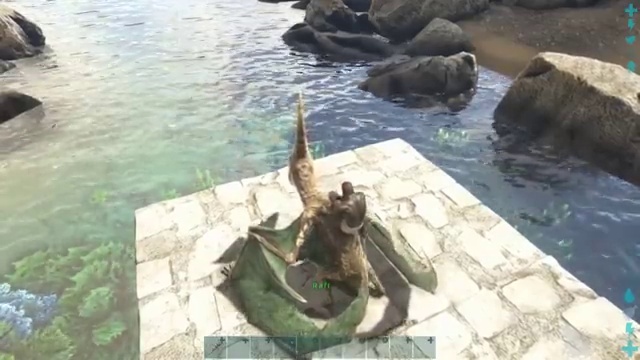'} & &
        \begin{minipage}[t]{\linewidth}\vspace{-15pt}
            \begin{minipage}[t]{0.8\linewidth}
                \textbf{Retrieved:} the next section which is surmounted by four male figures depicting the Navy, Infantry, Cavalry, and Artillery branches of the United States Army. Four female allegorical figures, ...
            \end{minipage}\hspace{1em}
            \begin{minipage}[t]{0.15\linewidth}
                \centering
                \raisebox{-26pt}[0pt][0pt]{\includegraphics[width=\linewidth]{figures/src/qualitative/casestudy/casestudy_image.jpg}}
            \end{minipage}
        \end{minipage}
        \\
        & \vspace{0.5pt} \textbf{Response:} Four people \redx & & \vspace{0.5pt} \textbf{Response:} Nine people \greencheck \\
        \bottomrule
    \end{tabular}
    \label{tab:casestudy}
    \vspace{-0.05in}
\end{table*}

%% file: sections/4_related_work.tex
\section{Related Work}

\paragraph{Large Vision Language Models}
Building on the impressive performance of LLMs~\citep{Gemini, gpt4o}, recent studies have extended them to visual domains. \citet{LLaVA} incorporates a CLIP-based~\citep{CLIP} image encoder to align visual inputs with language representations, followed by models using diverse encoders~\citep{Qwen-VL, InternVL, LLaVA1.5} and extensions to video~\citep{LLaVA-OneVision, internvl3_5, qwen3vl}. However, despite improved performance on multimodal benchmarks~\citep{DocVQA, MMMU, MVBench, video-mme} from larger datasets and with improved architectures, LVLMs still often suffer from hallucinations~\citep{HallucinationSurvey} when relying solely on parametric knowledge.

\paragraph{Retrieval-Augmented Generation}
To address the aforementioned limitation of parametric-only models, RAG incorporates external knowledge during response generation. While conventional RAG focuses on the textual corpus~\citep{RAG, RALM}, recent work extends it to multimodal sources such as images and videos~\citep{MuRAG, VideoRAG, ImageRAG}. However, these approaches assume a fixed single-modality retrieval, making them less adaptable to real-world queries that may require information from different modalities. Multimodal encoders~\citep{CLIP, GME, pecore, vlm2vecv2} enable unified embedding spaces across modalities, and \citet{unirag} retrieves from such spaces, but often fails to retrieve visual content for text queries. RAG-Anything~\citep{rag-anything} sidesteps this by converting all the multimodal knowledge into textual form, at the cost of heavy preprocessing and loss of modality-specific information. Other approaches~\citep{more, hmrag} retrieve from all modalities, followed by extra selection mechanisms, incurring notable computational cost. Lastly, adaptive retrieval strategies~\citep{Adaptive-RAG, Open-RAG, Rowen, Seakr, MBA-RAG} address query diversity but remain restricted to a single corpus~\citep{RetrievalQA, Dyn-VQA}.

\paragraph{Retrieval Granularity}
While most of the existing RAG methods operate at fixed granularity (e.g., full documents, passages, or sentences), real-world queries often require information at varying levels of specificity depending on the knowledge needed, which in turn impacts performance and efficiency in both textual~\citep{DenseXRetireval, LGMGC, MoG} and video-based retrieval systems~\citep{JSG}. In contrast, UniversalRAG performs query-level routing across modality and granularity dimensions, enabling retrieval from the most relevant source at the appropriate level.

%% file: sections/5_conclusion.tex
\section{Conclusion}
\label{sec:conclusion}

In this paper, we proposed UniversalRAG, a novel RAG framework designed to retrieve from corpora of diverse modalities and granularities. Through a modality- and granularity-aware routing mechanism, UniversalRAG dynamically selects the most suitable knowledge sources for each query, effectively addressing the limitations posed by modality gaps and fixed-granularity retrieval, which we further justify through theoretical results. Empirical evaluations across 10 benchmarks demonstrate that UniversalRAG outperforms both modality-specific and unified baselines, showcasing robust performance across diverse modalities. Also, our analyses highlight the importance of fine-grained retrieval and the complementary strengths of training-free and trained routers. We believe these findings demonstrate the potential of UniversalRAG as an adaptive solution for grounding LVLMs with heterogeneous external knowledge, paving the way for the one-for-all RAG that unifies the fragmented landscape of existing corpus-specific RAGs.

%% file: sections/x_limitation.tex
\section*{Limitations}
\label{sec:limitation}

The proposed UniversalRAG is designed for leveraging heterogeneous, multimodal corpora at RAG, enabling corpus-aware routing to flexibly utilize modality- and granularity-specific corpora. It is worth noting that the routing mechanism is its central part, and to improve its accuracy, high-quality samples for training may be required; however, existing datasets or benchmarks lack ground-truth labels indicating ideal modality or granularity for each query. Nonetheless, we address this by automatically annotating queries (based on inductive biases inherent in datasets or downstream performance measured with all the available corpora), as detailed in \cref{sec:appendix_dataset}. However, since they may contain some noise, constructing high-quality, human-annotated routing datasets would be a valuable direction for future work. Also, due to similar reasons: the absence of annotated data (specifically, the query-granularity pairs), we segment each (text and video) modality into two levels of granularity to obtain supervision signals for router training. Again, collecting more fine-grained annotations that cover a wider range of query-modality and query-granularity pairs would be an exciting direction to expand the applicability of UniversalRAG.

%% file: sections/x_ethics.tex
\section*{Ethical Considerations}

The proposed UniversalRAG can be seamlessly integrated with any LVLMs and compatible retrieval corpora, reducing hallucination with the corpus-specific routing. However, there can be potential private, harmful, or biased content present in the retrieved or generated outputs, depending on the nature of the underlying corpora or the internalized knowledge within LVLMs. To mitigate such risks, it is recommended to apply safeguard mechanisms and filtering techniques in retrieval and generation, to ensure the safe and responsible deployment.

%% file: sections/x_acknowledgement.tex
\section*{Acknowledgements}

This work was supported by the Institute for Information \& communications Technology Technology Planning \& Evaluation (IITP) grant funded by the Korea government (MSIT) (RS-2019-II190075, Artificial Intelligence Graduate School Program (KAIST)), the National Research Foundation of Korea (NRF) grant funded by the Korea government (MSIT) (RS-2023-00256259 \& RS-2026-25488933), the InnoCORE program of the Ministry of Science and ICT (No. N10250156), the Korea Machine Learning Ledger Orchestration for Drug Discovery Project (K-MELLODDY) grant funded by the Ministry of Health \& Welfare and Ministry of Science and ICT, Republic of Korea (RS-2024-00460870), the Institute of Information \& communications Technology Planning \& Evaluation (IITP) grant funded by the Korea government (MSIT) (RS-2022-II220713, Meta-learning Applicable to Real-world Problems), and the Center for Applied Research in Artificial Intelligence (CARAI) grant funded by DAPA and ADD (UD190031RD).

%% file: sections/x_appendix.tex
\clearpage
\appendix

\section{Additional Details on Dataset}
\label{sec:appendix_dataset}

\cref{tab:dataset_summary} provides an overview of all datasets and their corresponding knowledge corpora used in our experiments, including the target modality type as well as the size of the queries and corpora. We divide each dataset into a 3:7 ratio for training and testing. We offer the details of each dataset below.

\subsection{In-Domain Dataset}

\paragraph{MMLU} As a dataset comprising queries that can be answered without the need for retrieval, we use MMLU~\citep{MMLU}, a benchmark that spans a wide range of tasks, including problem-solving abilities (e.g., elementary mathematics, computer science) and world knowledge (e.g., law, world religions). Specifically, we use questions from all tasks in the development split.

\paragraph{Natural Questions (NQ)} We also use Natural Questions~\citep{nq}, a question answering dataset consisting of real user queries issued to the Google search engine, with answers annotated based on supporting Wikipedia articles. We randomly sample 2,000 QA pairs from the dev split, and formulate the text corpus by segmenting the Wikipedia corpus into paragraphs of at most 100 words.

\paragraph{HotpotQA} HotpotQA~\citep{hotpotqa} is a Wikipedia-based QA benchmark, but it contains complex queries that are annotated to reason over multiple articles. We utilize 2,000 randomly sampled QA pairs of the test split. As it requires multi-hop reasoning over multiple documents, we formulate the text corpus by grouping multiple related documents following LongRAG~\citep{longrag}, which can be longer than 4K tokens.

\paragraph{HybridQA} HybridQA~\citep{hybridqa} is a benchmark that requires reasoning over both tabular and textual information. Each question is grounded in a Wikipedia table, but often requires linking to associated text information to locate the correct answer. We randomly sample 2,000 QA pairs from the dev split. Unlike the original benchmark, which directly connects tables and textual evidence, we separate them into distinct table and text corpora to better validate our modality-specific routing-based retrieval framework.

\paragraph{MRAG-Bench (MRAG)} We utilize MRAG-Bench~\citep{mrag-bench}, a vision-centric RAG benchmark that requires only relevant images and does not rely on other modalities, and evaluate on all 1,353 questions. Unlike conventional text-only queries, each query in MRAG-Bench is multimodal, consisting of a textual question interleaved with a query image. We construct a single image corpus by collecting all images across questions.

\paragraph{WebQA} WebQA~\citep{WebQA} is a benchmark designed to evaluate the ability of LVLMs to reason over multiple sources of information, including both text and images, in an open-domain setting. As the dataset is originally constructed with question-specific retrieval sources that combine text and images, we extract a subset of questions that require retrieval of image for answering. We then further filter these using GPT-4o~\citep{gpt4o} with the prompt shown in \cref{fig:prompt_filter} to make sure questions are not grounded to a certain image, resulting in a final set of 2,000 QA pairs. Finally, we construct separate text and image corpora by extracting and aggregating evidence from each modality.

\paragraph{InfoSeek} InfoSeek~\citep{infoseek} is an open-domain benchmark comprising questions interleaved with images, which are best answered by retrieving relevant textual and visual information. For our experiments, we sample 2,000 QA pairs from the dev split and collect the text and image evidence associated with each question to construct the corresponding text and image corpora.

\input{tables/tab_dataset}

\paragraph{LVBench} LVBench~\citep{lvbench} is a benchmark developed for long video understanding, featuring questions generated by annotators based on YouTube videos with an average duration of over one hour. Since the benchmark was originally designed for non-RAG tasks, we rephrase the original text-video interleaved queries into a text-only format to align with our experimental setup using GPT-4o, with video metadata and a prompt (\cref{fig:prompt_rephrase}). Each query is associated with a specific video and a corresponding time range. Notably, the majority of queries are annotated with timestamps spanning less than five minutes, thereby focusing on short segments within the longer videos. Since some videos are currently unavailable, we conduct our evaluation on the available videos and their corresponding questions. For training, we use these short-timestamp queries as a clip-level dataset.

\paragraph{VideoRAG} We also utilize VideoRAG-Wiki and VideoRAG-Synth benchmarks, introduced in VideoRAG~\citep{VideoRAG}, which are designed to evaluate RAG over a video corpus. These benchmarks are built on the HowTo100M~\citep{howto100m} corpus (a large-scale collection of instructional YouTube videos) with queries sourced from WikiHowQA~\citep{wikihowqa} and synthetically generated QA pairs based on the videos. Since they lack timestamp annotations, we employ GPT-4o to identify video-level queries that are better answered through full video retrieval rather than short segments from the ground-truth video, which are then used as a video-level dataset for training the router.

\subsection{Out-of-Domain Dataset}
\label{sec:appendix_dataset_ood}

Unlike the in-domain datasets, the out-of-domain datasets are used solely for evaluation to assess the generalizability of our routing approach and consist only of test splits.

\paragraph{TruthfulQA} TruthfulQA~\citep{truthfulqa} includes general knowledge questions designed to test whether LLMs can avoid common false beliefs or misconceptions, on diverse categories, including health, law, and politics. We use the multiple-choice version of the dataset, which includes only a single correct answer per question.

\paragraph{TriviaQA} TriviaQA~\citep{triviaqa} is a reading comprehension dataset consisting of trivia questions paired with evidence texts sourced from Wikipedia and the web. To distinguish between queries that require text retrieval and those that do not, we categorize each query based on whether GPT-4o can produce an exact-match answer without access to external text. We randomly sample QA pairs from the dev split. Following the preprocessing strategies used in SQuAD and NQ, all supporting evidence documents are segmented into paragraphs of no more than 100 words.

\paragraph{SQuAD} SQuAD v1.1~\cite{squad} is a benchmark dataset consisting of questions generated by crowdworkers based on a set of Wikipedia articles. Each question is answerable given the appropriate context paragraph. From the dataset's 100,000+ QA pairs, we randomly sample 2,000 pairs of dev split. For context retrieval, we utilize the full provided Wikipedia corpus, segmenting each article into paragraphs of at most 100 words.

\paragraph{2WikiMultiHopQA} We also utilize 2WikiMultiHopQA~\citep{2wikimultihopqa}, a benchmark designed to evaluate multi-hop reasoning across two Wikipedia articles. We randomly sample 2,000 QA pairs from the dev split and construct a document-level corpus by aggregating all annotated candidate paragraph-level contexts for each question.

\paragraph{Visual-RAG} Visual-RAG~\citep{visualrag} is a question-answering benchmark designed for visual knowledge-intensive questions, specifically tailored for text-to-image retrieval tasks. We utilize the full set of provided queries but sample five images per category to construct the image retrieval pool, ensuring efficient text-to-image retrieval.

\paragraph{CinePile} CinePile~\citep{cinepile} is a long-video question-answering benchmark that features questions based on movie clips from YouTube. Since the benchmark was originally designed for video understanding tasks rather than RAG, we reformulate each query using the same procedure as LVBench. For each of the 144 available videos, we randomly select 10 questions from the test split. Since CinePile does not provide granularity annotations, we classify the questions into two categories (such as clip-level and full-video-level granularity) using GPT-4o, following the same approach used in VideoRAG.

\subsection{Evaluation Metrics}

We report results with standard metrics. For datasets with multiple-choice questions, we report Top-1 Accuracy (Acc), the proportion of questions answered correctly. For short-answer datasets, we use Exact Match (EM) and F1, which respectively measure exact agreement and word-level overlap between predictions and references; for InfoSeek, we use the custom accuracy metric defined in the original paper and official repository. For datasets with longer free-form answers, we use ROUGE-L, which captures the longest common subsequences between the prediction and reference~\citep{ROUGE}, and BERTScore, which assesses their semantic similarity~\citep{BERTScore}. We report the average score by averaging first within each modality, then across modalities. Results are obtained from a single run under limited computational resources, while we validate the generality of our framework across multiple backbone models.

\section{Additional Implementation Details}
\label{sec:appendix_implementation}

To effectively leverage both visual and textual information for visual element retrieval, we employ an ensemble approach that combines visual and textual similarity scores with a weighting ratio of 0.8 for visual information. The textual information consists of image captions for images and scripts for videos.
To handle long videos, we utilize PySceneDetect~\citep{PySceneDetect}, an open-source tool that detects scene boundaries by analyzing content changes (e.g., color histogram differences or threshold-based detection), to segment long videos into shorter clips with an average length of no more than 3 minutes.
Moreover, for both the retrieval and generation stages, we uniformly sample 32 frames per video. For baseline models that do not natively support video input, specifically UniRAG (which utilizes CLIP) and GME, we average the embeddings of these sampled frames to obtain a single representative embedding vector.

Training-based routers employ a lightweight classifier head on top of the backbone model to produce logits over multi-label prediction. Multi-label targets are converted into multi-hot vectors, and training is performed via binary cross-entropy loss between these targets and the predicted logits. The router is trained for 5 epochs with a learning rate of 2e-5 and a LoRA rank of $r=32$. At inference time, routing decisions are made using a predefined threshold of 0.8, selecting all modality-granularity combinations whose sigmoid probabilities exceed the threshold. In contrast, for the training-free variant, we prompt the model using a curated prompt that specifies task objectives and few-shot examples, as shown in \cref{fig:prompt_route}. Most experiments are conducted on NVIDIA RTX Pro 6000 Max-Q GPUs with 96GB of VRAM.

\section{Theoretical Analyses of UniversalRAG}
\label{sec:appendix_prop}

In this section, we present formal analyses of each module in UniversalRAG, including the effectiveness of modality routing (\cref{sec:appendix_theory_modality}) and multigranularity (\cref{sec:appendix_theory_granularity}), as well as the efficiency of modality-aware routing (\cref{sec:appendix_theory_efficiency}).

\subsection{Effectiveness of Modality Routing}
\label{sec:appendix_theory_modality}

For a rigorous analysis of the effectiveness of modality routing, we restate \cref{prop:unifiedemb} and provide a complete proof.

\begin{proposition}
    Let the similarity score in a unified embedding space $\mathcal{C}_{\text{\normalfont\texttt{unified}}}$ be defined as
    \[
        s(\vq, \vc)=\alpha\cdot \mathbf{1}\{m(\vq)=m(\vc)\}+\beta\cdot r(\vq, \vc)+\varepsilon,
    \]
    where $\alpha>0$ is a modality bias, $m(\cdot)$ denotes the modality, and $r(\cdot, \cdot)$ measures semantic relevance. If $\alpha$ is sufficiently large relative to the variance of $r$, the probability of retrieving items from the required modality $m^\ast(\vq)$ is less than under modality-aware routing followed by within-modality retrieval.
\end{proposition}

\begin{proof}
    Without loss of generality, let us consider the top-1 retrieval, as the extension to the top-$k$ case follows directly. Let the unified retrieval corpus $\mathcal{C}_\texttt{unified}$ be decomposed into three disjoint sets:
    \begin{equation}
    \begin{aligned}
        & S=\{\vc:\, m(\vc)=m(\vq)\}\\
        & R=\{\vc:\, m(\vc)=m^\ast(\vq)\}\\
        & O=\mathcal{C}_{\texttt{unified}}\setminus(S\cup R).
    \end{aligned}
    \end{equation}
    Let us consider the scenario where $m^\ast (\vq)\neq m(\vq)$ and $S, R\neq \emptyset$. Define $X_c\coloneq \beta \cdot r(\vq, \vc)+\varepsilon_\vc$ and suppose $\{X_\vc\}_{\vc\in \mathcal{C}_\texttt{unified}}$ are independent, mean-zero, sub-Gaussian with variance proxy $\sigma^2=\beta^2 \cdot\text{Var}[r(\vq, \vc)]+\text{Var}[\varepsilon_\vc]$. Then the similarity scores can be expressed as
    \begin{equation}
        s(\vq, \vc)=\begin{cases}
            \alpha + X_\vc, & \vc\in S \\
            X_\vc, & \vc\in R \cup O.
        \end{cases}
    \end{equation}
    Let $M_S = \max_{\vs\in S} X_\vs$, $M_R = \max_{\vr\in R} X_\vr$, and $M_O = \max_{\vo\in O}  X_\vo$. Under the unified embedding retrieval, the top-1 item lies in $R$ if and only if
    \[
        M_R \geq \alpha + \max\{M_S, M_O\}.
    \]
    Hence, we can obtain the upper bound of the probability where top-1 retrieval comes from $R$:
    \begin{equation}
    \begin{aligned}
        \mathbb{P}(\mathcal{T}_\texttt{unified}(\vq; \mathcal{C}_\texttt{unified})\in R)
        & = \mathbb{P}(M_R \geq \alpha + \max\{M_S, M_O\} \\
        & \leq \mathbb{P}(M_R-M_S \geq \alpha).
    \end{aligned}
    \label{eq:prop_unifiedemb-1}
    \end{equation}
    As $\{M_R-M_S\geq \alpha\}\subseteq\cup_{(\vr, \vs)\in R\times S}\{X_\vr - X_\vs \geq \alpha\}$, by the union bound we have
    \begin{equation}
        \mathbb{P}(M_R-M_S \geq \alpha) \leq \sum_{(\vr, \vs)\in R\times S}\mathbb{P}(X_\vr - X_\vs \geq \alpha).
    \end{equation}
    As $X_\vr - X_\vs$ is sub-Gaussian with variance proxy $2\sigma^2$, the Chernoff bound of the tail probability combined with \cref{eq:prop_unifiedemb-1} leads to
    \begin{equation}
        \mathbb{P}(\mathcal{T}_\texttt{unified}(\vq; \mathcal{C}_\texttt{unified})\in R) \leq |R| |S| \exp\left(-\frac{\alpha^2}{4\sigma^2}\right).
    \label{eq:prop_unifiedemb-2}
    \end{equation}
    By contrast, if the retrieval is done at the modality-specific corpus after modality-aware routing with accuracy $r$, the probability where the top-1 item is in $R$ is $r$. Combining this with \cref{eq:prop_unifiedemb-2},
    \begin{equation}
    \begin{aligned}
        \mathbb{P}(\mathcal{T}_\texttt{unified}(\vq; \mathcal{C}_\texttt{unified})\in R)
        & \leq |R| |S| \exp\left(-\frac{\alpha^2}{4\sigma^2}\right) \\
        & < r = \mathbb{P}(\mathcal{T}_{\mathcal{R}(\vq)}(\vq; \mathcal{C}_{\mathcal{R}(\vq)})\in R)
    \end{aligned}
    \end{equation}
    whenever $\alpha>2\sigma\sqrt{\frac{\log(|R||S|)}{r}}$. Meanwhile, the right-hand side of \cref{eq:prop_unifiedemb-2} decays to 0 as $\alpha/\sigma \to \infty$. Hence, for $\alpha$ large enough relative to the variance of $r$, unified embedding retrieval is strictly worse than retrieving from modality-specific corpus after modality-aware routing.
\end{proof}

\begin{remark}
    Consider very large corpora with $|R|=|S|=10^{12}$. In this setting, if $p=0.8$ and $\sigma=0.01$, then $\alpha>2\sigma\sqrt{\frac{\log(|R||S|)}{p}}\simeq 0.17$ is sufficient to ensure that routing-based retrieval outperforms unified embedding retrieval. Given that most multimodal encoders exhibit inherent modality biases (as illustrated in \cref{fig:embedding,fig:appendix_embedding}), this underscores the necessity of modality-aware routing.
\end{remark}

\subsection{Effectiveness of Multigranularity}
\label{sec:appendix_theory_granularity}

In \cref{sec:experimental_results,sec:appendix_multigranularity}, we show that routing with multiple granularities within each modality improves performance (see \cref{tab:granularity_perf,tab:appendix_granularity}). We also provide a simple statement and proof that support these empirical findings.

\begin{proposition}
    Let $F(Q; m,g)$ be the expected response quality when retrieving from modality $m$ using granularity $g$. If there exist queries $\vq_1, \vq_2$ and granularities $g_f, g_c$ such that $F(\vq_1; m, g_f) > F(\vq_1; m, g_c)$ and $F(\vq_2; m, g_c) > F(\vq_2; m, g_f)$, then a routing policy that assigns $g_f$ to $\vq_1$ and $g_c$ to $\vq_2$ attains strictly higher expected quality than any fixed-granularity policy.
    \label{prop:multigranularity}
\end{proposition}

\begin{proof}
    Consider any fixed policy that always uses a single granularity $g\in \{g_f, g_c\}$. If $g=g_f$, then we have
    \begin{equation}
        F(\vq_1; m, g_f)+F(\vq_2;m,g_f) < F(\vq_1; m,g_f)+F(\vq_2;m, g_c).
    \end{equation}
    Similarly, if $g=g_c$, then we have
    \begin{equation}
        F(\vq_1; m, g_c)+F(\vq_2;m,g_c) < F(\vq_1; m,g_f)+F(\vq_2;m, g_c).
    \end{equation}
    In both cases, the sum of response quality with the routing policy that applies $g_f$ to $\vq_1$ and $g_c$ to $\vq_2$ strictly exceeds that of any fixed granularity $g$.
\end{proof}

\subsection{Efficiency of Modality-Specific Retrieval}
\label{sec:appendix_theory_efficiency}

While the empirical results in \cref{sec:experimental_results} demonstrate the efficiency benefits of modality-aware routing (with latency trends shown in \cref{fig:retrieval_latency}), we provide a more rigorous analysis on its computational advantages. Let $N$ denote the size of each modality- and granularity-specific corpus, assuming uniform corpus sizes for simplicity, and let $k$ be the number of available routing choices (i.e., the number of modality-granularity pairs). Under a unified embedding approach, retrieval is performed over a single aggregated corpus of size $kN$, incurring a search cost that scales with the total corpus size. In contrast, UniversalRAG first performs lightweight routing to select the most relevant modality-granularity subset, and then conducts retrieval over only a small selected subset.

\begin{proposition}
    Let $T(m)$ denote the expected retrieval latency of a single query over a corpus of size $m$ under a fixed retrieval backend, and let the routing cost be a fixed constant $C$, independent of the number of available routing choices $k>1$. Then, UniversalRAG achieves lower latency than unified embedding space retrieval on large-scale corpora.
\end{proposition}

\begin{proof}
    Under unified embedding, all modality-granularity corpora are merged into a single index of size $kN$. Then, the expected per-query retrieval latency is $T_\texttt{unified}=T(kN)$. Under UniversalRAG, routing incurs a constant overhead $C$ and then retrieval is executed only on a small number of routed corpora. Assuming retrieval calls of selected corpus are executed in parallel, the end-to-end latency of whole retrieval process is $T_\texttt{routing}=C+T(N)$. Let us first consider the case of exact retrieval with embeddings, where the backend exhibits linear scaling $T(m)=\Theta(m)$, then we obtain
    \begin{equation}
        \frac{T_\texttt{unified}}{T_\texttt{routing}}\gtrsim\frac{kN}{N+C}=\frac{k}{1+C/N}.
    \end{equation}
    Taking $N\to \infty$ yields
    \begin{equation}
        \liminf_{N\to\infty} \frac{T_{\texttt{unified}}}{T_{\texttt{routing}}} = \Theta(k),
    \end{equation}
    resulting in a linear-in-$k$ speedup. Meanwhile, many modern retrieval systems adopt approximate nearest neighbor search~\citep{faiss}, which can achieve logarithmic query-time scaling $T(m)=\Theta(\log m)$ (in the best case). Then, for sufficiently large $N$,
    \begin{equation}
        \frac{T_\texttt{unified}}{T_\texttt{routing}}\gtrsim\frac{\log (kN)}{\log N+C}=\frac{\log N + \log k}{\log N+C}.
    \end{equation}
    Letting $N\to \infty$, we have
    \begin{equation}
        \liminf_{N\to\infty} \frac{T_{\texttt{unified}}}{T_{\texttt{routing}}} \ge 1.
    \end{equation}
    Thus, even with the approximate retrieval with logarithmic scaling, UniversalRAG achieves a constant-factor asymptotic speedup. Combining these results, UniversalRAG attains strictly lower asymptotic retrieval latency than unified embedding space retrieval for any retrieval methods.
\end{proof}

\input{tables/tab_main_diverse_lvlms}

\clearpage
\section{Additional Experimental Results}

\subsection{Additional Results using Different LVLMs}

\cref{tab:appendix_diverse_lvlms} shows detailed generation results of baselines and UniversalRAG models on 10 benchmarks using InternVL3.5-8B and Molmo2-4B as generation models. In both settings, UniversalRAG outperforms all baselines and achieves average scores comparable to Oracle. These results demonstrate that UniversalRAG is robust and generalizable in various LVLM generators.

\subsection{Additional Results on Multigranularity}
\label{sec:appendix_multigranularity}

\cref{tab:granularity_perf} demonstrates the correlation between the number of granularity levels and end-to-end performance for two training-free models, leveraging the flexibility of our approach in scenarios without labeled data. We further extend this analysis to training-based routers, comparing performance with and without granularity. \cref{tab:appendix_granularity} reports results across three training-based router models, consistently demonstrating a performance advantage when granularity is incorporated. These findings underscore the efficacy of including granularity in routing decisions for both training-free and training-based approaches.

\input{tables/tab_appendix_granularity}
\input{tables/tab_main_ood}

\subsection{Detailed Results on Out-of-Domain Dataset}

We provide the generation results of UniversalRAG variants and baseline methods on each out-of-domain dataset in \cref{tab:appendix_main_ood}. Overall, UniversalRAG consistently outperforms all baselines on average. Notably, the training-free router variants exhibit strong performance across all datasets, showing their outstanding generalization ability to unseen queries. In contrast, trained routers achieve relatively lower performance than on in-domain datasets; nevertheless, they remain robust and still surpass the baseline methods by a large margin.

\input{figures/fig_appendix_embedding}

\section{Modality Gap in Unified Embedding Space}

\cref{fig:appendix_embedding} visualizes the modality gap within the unified embedding space of six multimodal encoders~\citep{e5-v, pecore, mm-embed, GME, vlm2vecv2, qwen3vlembedding}. The PCA plot reveals that embeddings cluster by modality, with text embeddings (shown in green) exhibiting larger distances from those of other modalities. Recent methods like E5-V, GME, and Qwen3-VL-Embedding focus on better aligning these modalities to narrow the gap. However, despite these efforts, a noticeable separation between modalities remains, indicating that current multimodal encoders still struggle to fully unify the embedding space across text, images, and videos. Therefore, the modality routing mechanism of UniversalRAG is required to dynamically direct each query to its corresponding modality-specific embedding space, thereby effectively bridging the modality gap and enhancing retrieval performance.

\section{Qualitative Results}
\label{sec:qualitative_results}

We present case studies to demonstrate the effectiveness of UniversalRAG. \cref{tab:qualitative_full} compares the results of various RAG approaches, including traditional single-modality methods and UniversalRAG, on queries from the WebQA dataset. Traditional approaches such as TextRAG and VideoRAG fail to generate accurate answers: TextRAG retrieves passages lacking relevant visual details, while VideoRAG is better suited for temporal reasoning tasks. In contrast, UniversalRAG correctly routes the query to the image modality, recognizing that visual information about color is necessary, and successfully generates the correct response. This highlights the advantage of modality-aware routing in leveraging the appropriate data from the correct modality corpus, demonstrating UniversalRAG's ability to adaptively select the most informative modalities and granularities for accurate answer generation.

In addition to modality routing, we observe that UniversalRAG also benefits from retrieving information at the appropriate granularity. \cref{tab:qualitative_gran_text} shows results from HotpotQA, where the query requires complex reasoning over multiple text sources. While paragraph-level granularity fails to provide sufficient context for reasoning, UniversalRAG routes the query to the document-level corpus to retrieve all the textual information necessary for accurate reasoning. Similarly, for video queries, \cref{tab:qualitative_gran_video} shows results from LVBench on the query that requires only a short segment of the full long video to answer. While full-video-level retrieval includes irrelevant content and uniformly sampled frames fail to capture the necessary information, clip-level retrieval focuses on smaller, more relevant segments of the video to ensure that only the most pertinent visual details are considered, leading to a more accurate answer.

UniversalRAG performs cross-modal retrieval, allowing the router to select multiple modality-granularity combinations when required, rather than restricting routing to a single source. \cref{tab:qualitative_cross} presents an example from HybridQA, where queries primarily rely on tabular data but benefit substantially from complementary textual evidence. In such cases, factual information is best captured from paragraphs, whereas structured knowledge, such as numerical values, is more effectively represented in tables. By jointly retrieving from both modalities, UniversalRAG effectively aggregates complementary evidence and provides the information necessary to answer the query correctly. In contrast, a unimodal variant that restricts retrieval to a single modality retrieves incomplete evidence and fails to support correct reasoning.

However, there are some cases where the routing mechanism fails, particularly when the query exhibits ambiguity in modality requirement or when the required information spans across multiple modalities. \cref{tab:qualitative_failure} shows failure cases in which UniversalRAG, employing GPT-5 as a training-free router, incorrectly routes the modality. In the first example, the router's prediction deviates from the inductive ground-truth label as GPT-5, as a modern frontier model, has prior knowledge beyond the predefined routing taxonomy. Although this results in a nominal misclassification, it does not affect the final generation quality, as the model can answer the query without external retrieval. The router also struggles to distinguish between closely related modalities. As illustrated in the second case, a query requiring temporally localized visual evidence is incorrectly routed from clip-level retrieval to static image retrieval. Moreover, the router sometimes exhibits difficulty in determining the appropriate retrieval granularity. Queries that lie near the boundary between different granularity levels are sometimes misrouted, as shown in the third and fourth examples. Finally, the router occasionally fails to recognize cross-modal information needs, leading to incorrect routing decisions for queries that require joint reasoning across modalities, as illustrated in the last example.

\clearpage
\input{tables/tab_qualitative_full}
\input{tables/tab_qualitative_gran_text}
\input{tables/tab_qualitative_gran_video}
\input{tables/tab_qualitative_cross_modal}
\input{tables/tab_qualitative_failure}
\input{figures/fig_prompts}

%% file: tables/tab_dataset.tex
\begin{table*}[t]
    \centering
    \caption{Dataset summary for in-domain and out-of-domain benchmarks. Average corpus length denotes the mean token count for text corpora and the mean duration for video corpora.}
    \resizebox{0.8\linewidth}{!}{
        \begin{tabular}{l c c c c c}
            \toprule
            \textbf{Dataset} & \textbf{Query Modality} & \textbf{Target Retrieval Modality} & \textbf{\# Queries} & \textbf{Corpus Size} & \textbf{Avg Length} \\
            \midrule
            \midrule
            \rowcolor{blue!5}\multicolumn{6}{c}{\textit{In-Domain Datasets}} \\
            \midrule
            MMLU & Text & None & 1,710 & - & - \\
            Natural Questions & Text & Paragraph & 2,000 & 850k & 100 tokens \\
            HotpotQA & Text & Document & 2,000 & 509k & 693 tokens \\
            HybridQA & Text & Paragraph + Table & 2,000 & 15k & - \\
            MRAG-Bench & Text + Image & Image & 1,353 & 6k & - \\
            WebQA & Text & Paragraph + Image & 2,000 & 20k & - \\
            InfoSeek & Text + Image & Paragraph + Image & 2,000 & 20k & - \\
            LVBench & Text & Clip/Video & 777 & 89 & 3,865s \\
            VideoRAG-Wiki & Text & Clip/Video & 374 & \multirow{2}{*}{9k} & \multirow{2}{*}{378s} \\
            VideoRAG-Synth & Text & Clip/Video & 374 & & \\
            \midrule
            \midrule
            \rowcolor{blue!5}\multicolumn{6}{c}{\textit{Out-of-Domain Datasets}} \\
            \midrule
            TruthfulQA & Text & None & 790 & - & - \\
            TriviaQA & Text & Paragraph & 661 & 661k & 100 tokens \\
            SQuAD & Text & Paragraph & 2,000 & 1.19M & 100 tokens \\
            2WikiMultiHopQA & Text & Document & 2,000 & 12k & 562 tokens \\
            Visual-RAG & Text & Image & 374 & 2k & - \\
            CinePile & Text & Clip/Video & 1,440 & 144 & 158s \\
            \bottomrule
        \end{tabular}
    }
    \label{tab:dataset_summary}
\end{table*}

%% file: tables/tab_main_diverse_lvlms.tex
\begin{table*}[t]
    \centering
    \caption{Results of diverse RAG methods with diverse LVLMs (InternVL3.5-8B and Molmo2-4B) across modalities. \textbf{Bold} denotes the best performance and \underline{underlined} indicates the second-best among UniversalRAG variants, using either \textcolor{teal!80}{trained} or \textcolor{blue!80}{training-free} routers. R-L and BERT correspond to ROUGE-L and BERTScore, respectively.}
    \renewcommand{\tabcolsep}{1.6mm}
    \small
    \resizebox{\textwidth}{!}{
        \begin{tabular}{l l c cc cc cc c cc c c cc cc c}
            \toprule
            \multicolumn{2}{c}{} & \textbf{MMLU} & \multicolumn{2}{c}{\textbf{NQ}} & \multicolumn{2}{c}{\textbf{HotpotQA}} & \multicolumn{2}{c}{\textbf{HybridQA}} & \textbf{MRAG} & \multicolumn{2}{c}{\textbf{WebQA}} & \textbf{InfoSeek} & \textbf{LVBench} & \multicolumn{2}{c}{\textbf{VideoRAG-Wiki}} & \multicolumn{2}{c}{\textbf{VideoRAG-Synth}} & \multirow[c]{2}{*}[-0.3em]{\textbf{Avg}} \\
            \cmidrule(lr){3-3} \cmidrule(lr){4-5} \cmidrule(lr){6-7} \cmidrule(lr){8-9} \cmidrule(lr){10-10} \cmidrule(lr){11-12} \cmidrule(lr){13-13} \cmidrule(lr){14-14} \cmidrule(lr){15-16} \cmidrule(lr){17-18}
            & \textbf{Models} & Acc & EM & F1 & EM & F1 & EM & F1 & Acc & R-L & BERT & Acc & Acc & R-L & BERT & R-L & BERT \\
            \midrule
            \midrule
            \multirow{21}{*}[-1.0em]{\rotatebox[origin=c]{90}{\textbf{InternVL3.5-8B}}} & Naïve & 71.58 & 11.75 & 20.59 & 14.85 & 22.02 & 1.60 & 5.15 & 42.50 & 56.95 & 93.64 & 8.05 & 28.31 & 20.90 & 87.39 & 34.41 & 90.52 & 31.29 \\
            \noalign{\vskip 0.25ex}\cdashline{2-19}\noalign{\vskip 0.75ex}
            & ParagraphRAG & 68.48 & 33.60 & 46.05 & 19.20 & 26.27 & 4.25 & 7.69 & 36.81 & 34.15 & 89.61 & 13.25 & 22.52 & 17.62 & 85.52 & 27.08 & 88.80 & 31.73 \\
            & DocumentRAG & 69.30 & 19.40 & 26.85 & 24.90 & 33.40 & 3.35 & 7.37 & 35.03 & 34.36 & 89.54 & 11.25 & 29.60 & 16.37 & 84.86 & 24.07 & 88.04 & 30.57 \\
            & TableRAG & 63.22 & 6.05 & 9.85 & 11.80 & 16.47 & 7.30 & 11.31 & 40.06 & 28.99 & 88.59 & 4.10 & 26.38 & 14.27 & 83.78 & 21.22 & 86.97 & 25.20 \\
            & ImageRAG & 72.75 & 11.65 & 18.79 & 14.85 & 21.62 & 1.75 & 5.09 & 47.89 & 58.97 & 93.85 & 11.15 & 29.21 & 20.97 & 87.50 & 34.77 & 90.55 & 32.29 \\
            & ClipRAG & 69.94 & 9.25 & 15.00 & 12.60 & 18.38 & 1.95 & 4.24 & 32.82 & 14.48 & 85.00 & 6.00 & 36.04 & 21.68 & 88.09 & 35.43 & 90.93 & 26.90 \\
            & VideoRAG & 70.29 & 10.10 & 16.08 & 14.30 & 19.53 & 1.30 & 3.97 & 33.48 & 14.07 & 84.57 & 5.35 & 35.78 & 22.17 & 89.14 & 36.97 & 91.47 & 27.30 \\
            \noalign{\vskip 0.25ex}\cdashline{2-19}\noalign{\vskip 0.75ex}
            & UniRAG & 69.65 & 14.85 & 23.82 & 17.40 & 25.34 & 2.85 & 6.78 & 34.96 & 34.38 & 89.77 & 10.45 & 23.68 & 18.31 & 86.02 & 25.93 & 88.55 & 28.60 \\
            & GME & 69.18 & 15.40 & 24.53 & 17.15 & 25.31 & 2.60 & 6.59 & 35.33 & 34.22 & 89.73 & 11.10 & 23.42 & 17.23 & 85.39 & 25.13 & 88.41 & 28.53 \\
            & $\text{PE}_{\text{core}}$ & 69.24 & 14.90 & 23.91 & 17.50 & 25.74 & 2.75 & 6.65 & 34.81 & 31.74 & 89.02 & 10.70 & 24.07 & 17.68 & 85.50 & 25.16 & 88.32 & 28.38 \\
            & VLM2Vec-V2 & 69.65 & 15.25 & 24.35 & 16.75 & 24.89 & 3.15 & 7.14 & 35.70 & 32.05 & 89.23 & 10.85 & 23.04 & 17.41 & 85.42 & 26.42 & 88.71 & 28.52 \\
            \noalign{\vskip 0.25ex}\cdashline{2-19}\noalign{\vskip 0.75ex}
            & MultiRAG & 68.54 & 18.80 & 28.10 & 18.90 & 26.11 & 3.50 & 7.62 & 37.92 & 37.52 & 90.34 & 11.40 & 22.91 & 18.52 & 86.24 & 26.48 & 88.63 & 29.77 \\
            \noalign{\vskip 0.25ex}\cdashline{2-19}\noalign{\vskip 0.75ex}
            & \multicolumn{18}{l}{\textbf{UniversalRAG (Ours)}} \\
            \addlinespace[0.2ex]
            & \cellcolor{green!8}\hspace{1em}\textbf{\textit{Trained Routers}} & \cellcolor{green!8} & \cellcolor{green!8} & \cellcolor{green!8} & \cellcolor{green!8} & \cellcolor{green!8} & \cellcolor{green!8} & \cellcolor{green!8} & \cellcolor{green!8} & \cellcolor{green!8} & \cellcolor{green!8} & \cellcolor{green!8} & \cellcolor{green!8} & \cellcolor{green!8} & \cellcolor{green!8} & \cellcolor{green!8} & \cellcolor{green!8} & \cellcolor{green!8} \\
            & \cellcolor{green!8}\hspace{2.5em}Qwen3-VL-2B-Instruct & \cellcolor{green!8}\underline{71.58} & \cellcolor{green!8}\underline{33.25} & \cellcolor{green!8}\underline{45.58} & \cellcolor{green!8}\underline{24.50} & \cellcolor{green!8}\underline{33.07} & \cellcolor{green!8}\textbf{10.25} & \cellcolor{green!8}\textbf{14.52} & \cellcolor{green!8}\textbf{47.89} & \cellcolor{green!8}\textbf{61.34} & \cellcolor{green!8}\textbf{94.05} & \cellcolor{green!8}\textbf{15.95} & \cellcolor{green!8}\textbf{36.04} & \cellcolor{green!8}\textbf{22.02} & \cellcolor{green!8}\textbf{89.11} & \cellcolor{green!8}\textbf{36.92} & \cellcolor{green!8}\textbf{91.49} & \cellcolor{green!8}\textbf{39.60} \\
            & \cellcolor{green!8}\hspace{2.5em}InternVL3.5-1B & \cellcolor{green!8}\underline{71.58} & \cellcolor{green!8}33.10 & \cellcolor{green!8}45.27 & \cellcolor{green!8}\textbf{24.70} & \cellcolor{green!8}\textbf{33.19} & \cellcolor{green!8}10.05 & \cellcolor{green!8}14.28 & \cellcolor{green!8}\textbf{47.89} & \cellcolor{green!8}60.98 & \cellcolor{green!8}\underline{93.86} & \cellcolor{green!8}\underline{15.70} & \cellcolor{green!8}\textbf{36.04} & \cellcolor{green!8}21.96 & \cellcolor{green!8}\underline{88.97} & \cellcolor{green!8}\underline{36.79} & \cellcolor{green!8}\underline{91.43} & \cellcolor{green!8}\underline{39.50} \\
            & \cellcolor{green!8}\hspace{2.5em}T5Gemma 2 270M & \cellcolor{green!8}\textbf{71.87} & \cellcolor{green!8}\textbf{33.40} & \cellcolor{green!8}\textbf{45.70} & \cellcolor{green!8}24.15 & \cellcolor{green!8}32.29 & \cellcolor{green!8}\underline{10.10} & \cellcolor{green!8}\underline{14.33} & \cellcolor{green!8}46.71 & \cellcolor{green!8}\underline{61.04} & \cellcolor{green!8}93.75 & \cellcolor{green!8}\underline{15.70} & \cellcolor{green!8}35.26 & \cellcolor{green!8}\underline{22.00} & \cellcolor{green!8}88.94 & \cellcolor{green!8}36.74 & \cellcolor{green!8}91.38 & \cellcolor{green!8}39.24 \\
            \addlinespace[0.2ex]
            & \cellcolor{blue!8}\hspace{1em}\textbf{\textit{Training-free Routers}} & \cellcolor{blue!8} & \cellcolor{blue!8} & \cellcolor{blue!8} & \cellcolor{blue!8} & \cellcolor{blue!8} & \cellcolor{blue!8} & \cellcolor{blue!8} & \cellcolor{blue!8} & \cellcolor{blue!8} & \cellcolor{blue!8} & \cellcolor{blue!8} & \cellcolor{blue!8} & \cellcolor{blue!8} & \cellcolor{blue!8} & \cellcolor{blue!8} & \cellcolor{blue!8} & \cellcolor{blue!8} \\
            & \cellcolor{blue!8}\hspace{2.5em}GPT-5 & \cellcolor{blue!8}70.99 & \cellcolor{blue!8}31.35 & \cellcolor{blue!8}43.82 & \cellcolor{blue!8}21.90 & \cellcolor{blue!8}30.61 & \cellcolor{blue!8}6.65 & \cellcolor{blue!8}10.73 & \cellcolor{blue!8}45.90 & \cellcolor{blue!8}48.87 & \cellcolor{blue!8}92.16 & \cellcolor{blue!8}12.85 & \cellcolor{blue!8}33.85 & \cellcolor{blue!8}19.14 & \cellcolor{blue!8}87.15 & \cellcolor{blue!8}31.24 & \cellcolor{blue!8}89.27 & \cellcolor{blue!8}36.49 \\
            & \cellcolor{blue!8}\hspace{2.5em}Qwen3-VL-8B-Instruct & \cellcolor{blue!8}71.17 & \cellcolor{blue!8}31.30 & \cellcolor{blue!8}43.69 & \cellcolor{blue!8}22.85 & \cellcolor{blue!8}31.57 & \cellcolor{blue!8}6.50 & \cellcolor{blue!8}10.58 & \cellcolor{blue!8}45.53 & \cellcolor{blue!8}50.32 & \cellcolor{blue!8}93.73 & \cellcolor{blue!8}13.20 & \cellcolor{blue!8}34.49 & \cellcolor{blue!8}19.06 & \cellcolor{blue!8}86.94 & \cellcolor{blue!8}31.08 & \cellcolor{blue!8}89.11 & \cellcolor{blue!8}36.73 \\
            \noalign{\vskip 0.25ex}\cdashline{2-19}\noalign{\vskip 0.75ex}
            & Oracle & 71.58 & 33.60 & 46.05 & 24.90 & 33.40 & 10.35 & 15.17 & 47.89 & 61.56 & 94.20 & 15.85 & 36.04 & 22.17 & 89.14 & 36.97 & 91.47 & 39.80 \\

            \midrule
            \midrule
            \multirow{21}{*}[-1.1em]{\rotatebox[origin=c]{90}{\textbf{Molmo2-4B}}} & Naïve & 70.12 & 9.80 & 18.75 & 14.40 & 23.79 & 2.05 & 6.36 & 48.41 & 64.38 & 94.80 & 10.40 & 32.17 & 21.50 & 87.58 & 35.60 & 90.75 & 33.19 \\
            \noalign{\vskip 0.25ex}\cdashline{2-19}\noalign{\vskip 0.75ex}
            & ParagraphRAG & 68.36 & 38.65 & 50.53 & 22.00 & 29.59 & 5.20 & 9.83 & 39.54 & 63.28 & 94.26 & 15.85 & 30.12 & 16.88 & 85.75 & 32.31 & 89.77 & 36.52 \\
            & DocumentRAG & 68.42 & 20.50 & 28.45 & 25.50 & 34.51 & 3.95 & 8.38 & 40.28 & 63.17 & 94.32 & 13.20 & 33.72 & 16.49 & 85.39 & 32.05 & 89.66 & 34.53 \\
            & TableRAG & 67.31 & 8.70 & 14.25 & 15.00 & 21.39 & 8.55 & 13.59 & 42.79 & 61.61 & 94.52 & 6.10 & 31.53 & 14.34 & 84.63 & 32.88 & 89.96 & 31.04 \\
            & ImageRAG & 69.88 & 11.00 & 18.70 & 16.35 & 23.73 & 1.70 & 5.73 & 52.55 & 71.53 & 96.31 & 12.30 & 32.30 & 21.12 & 87.42 & 33.20 & 90.46 & 34.00 \\
            & ClipRAG & 69.36 & 9.55 & 16.59 & 15.15 & 22.29 & 1.95 & 5.50 & 30.67 & 66.42 & 94.98 & 8.45 & 38.48 & 21.62 & 87.27 & 35.77 & 90.82 & 31.13 \\
            & VideoRAG & 69.12 & 9.75 & 16.98 & 15.65 & 23.13 & 1.50 & 5.22 & 30.75 & 65.98 & 94.90 & 5.90 & 36.55 & 21.98 & 87.91 & 35.96 & 91.04 & 30.83 \\
            \noalign{\vskip 0.25ex}\cdashline{2-19}\noalign{\vskip 0.75ex}
            & UniRAG & 67.95 & 12.10 & 21.05 & 16.35 & 24.12 & 3.85 & 8.26 & 41.54 & 62.83 & 94.28 & 13.55 & 33.98 & 17.21 & 86.03 & 32.54 & 89.91 & 32.50 \\
            & GME & 68.13 & 12.45 & 21.32 & 16.40 & 24.35 & 3.70 & 8.14 & 41.17 & 63.07 & 94.31 & 13.20 & 33.59 & 17.04 & 85.91 & 32.18 & 89.73 & 32.43 \\
            & $\text{PE}_{\text{core}}$ & 68.25 & 13.35 & 21.28 & 16.75 & 24.23 & 3.50 & 7.92 & 41.32 & 63.02 & 94.22 & 13.45 & 32.43 & 16.89 & 85.73 & 32.22 & 89.81 & 32.28 \\
            & VLM2Vec-V2 & 68.01 & 12.20 & 21.07 & 16.45 & 24.06 & 3.60 & 8.03 & 40.28 & 62.79 & 94.09 & 12.85 & 32.05 & 17.17 & 85.99 & 32.84 & 90.01 & 32.04 \\
            \noalign{\vskip 0.25ex}\cdashline{2-19}\noalign{\vskip 0.75ex}
            & MultiRAG & 68.42 & 13.30 & 22.44 & 18.20 & 25.43 & 4.15 & 8.51 & 42.27 & 64.14 & 94.50 & 14.60 & 32.18 & 16.70 & 85.68 & 32.45 & 89.90 & 32.90 \\
            \noalign{\vskip 0.25ex}\cdashline{2-19}\noalign{\vskip 0.75ex}
            & \multicolumn{18}{l}{\textbf{UniversalRAG (Ours)}} \\
            \addlinespace[0.2ex]
            & \cellcolor{green!8}\hspace{1em}\textbf{\textit{Trained Routers}} & \cellcolor{green!8} & \cellcolor{green!8} & \cellcolor{green!8} & \cellcolor{green!8} & \cellcolor{green!8} & \cellcolor{green!8} & \cellcolor{green!8} & \cellcolor{green!8} & \cellcolor{green!8} & \cellcolor{green!8} & \cellcolor{green!8} & \cellcolor{green!8} & \cellcolor{green!8} & \cellcolor{green!8} & \cellcolor{green!8} & \cellcolor{green!8} & \cellcolor{green!8} \\
            & \cellcolor{green!8}\hspace{2.5em}Qwen3-VL-2B-Instruct & \cellcolor{green!8}\textbf{70.12} & \cellcolor{green!8}\textbf{37.95} & \cellcolor{green!8}\textbf{49.83} & \cellcolor{green!8}\textbf{25.35} & \cellcolor{green!8}\textbf{34.30} & \cellcolor{green!8}\textbf{10.30} & \cellcolor{green!8}\textbf{15.23} & \cellcolor{green!8}\underline{52.55} & \cellcolor{green!8}\textbf{73.38} & \cellcolor{green!8}\textbf{96.89} & \cellcolor{green!8}\underline{17.20} & \cellcolor{green!8}\textbf{38.61} & \cellcolor{green!8}\underline{21.72} & \cellcolor{green!8}\textbf{87.56} & \cellcolor{green!8}\textbf{35.68} & \cellcolor{green!8}\textbf{90.80} & \cellcolor{green!8}\textbf{41.83} \\
            & \cellcolor{green!8}\hspace{2.5em}InternVL3.5-1B & \cellcolor{green!8}\textbf{70.12} & \cellcolor{green!8}37.85 & \cellcolor{green!8}\underline{49.62} & \cellcolor{green!8}\textbf{25.35} & \cellcolor{green!8}\textbf{34.30} & \cellcolor{green!8}\underline{10.15} & \cellcolor{green!8}\underline{15.08} & \cellcolor{green!8}\underline{52.55} & \cellcolor{green!8}\underline{73.27} & \cellcolor{green!8}\underline{96.81} & \cellcolor{green!8}\textbf{17.35} & \cellcolor{green!8}\underline{38.48} & \cellcolor{green!8}\textbf{21.73} & \cellcolor{green!8}87.54 & \cellcolor{green!8}\underline{35.57} & \cellcolor{green!8}\underline{90.79} & \cellcolor{green!8}\underline{41.76} \\
            & \cellcolor{green!8}\hspace{2.5em}T5Gemma 2 270M & \cellcolor{green!8}69.94 & \cellcolor{green!8}\underline{37.90} & \cellcolor{green!8}49.60 & \cellcolor{green!8}25.30 & \cellcolor{green!8}34.04 & \cellcolor{green!8}9.45 & \cellcolor{green!8}14.70 & \cellcolor{green!8}\textbf{52.70} & \cellcolor{green!8}73.08 & \cellcolor{green!8}96.75 & \cellcolor{green!8}17.00 & \cellcolor{green!8}37.32 & \cellcolor{green!8}21.65 & \cellcolor{green!8}\underline{87.55} & \cellcolor{green!8}35.38 & \cellcolor{green!8}90.73 & \cellcolor{green!8}41.48 \\
            \addlinespace[0.2ex]
            & \cellcolor{blue!8}\hspace{1em}\textbf{\textit{Training-free Routers}} & \cellcolor{blue!8} & \cellcolor{blue!8} & \cellcolor{blue!8} & \cellcolor{blue!8} & \cellcolor{blue!8} & \cellcolor{blue!8} & \cellcolor{blue!8} & \cellcolor{blue!8} & \cellcolor{blue!8} & \cellcolor{blue!8} & \cellcolor{blue!8} & \cellcolor{blue!8} & \cellcolor{blue!8} & \cellcolor{blue!8} & \cellcolor{blue!8} & \cellcolor{blue!8} & \cellcolor{blue!8} \\
            & \cellcolor{blue!8}\hspace{2.5em}GPT-5 & \cellcolor{blue!8}69.88 & \cellcolor{blue!8}32.80 & \cellcolor{blue!8}44.67 & \cellcolor{blue!8}23.05 & \cellcolor{blue!8}32.78 & \cellcolor{blue!8}5.75 & \cellcolor{blue!8}10.34 & \cellcolor{blue!8}51.07 & \cellcolor{blue!8}70.43 & \cellcolor{blue!8}95.47 & \cellcolor{blue!8}16.85 & \cellcolor{blue!8}36.55 & \cellcolor{blue!8}19.86 & \cellcolor{blue!8}86.61 & \cellcolor{blue!8}33.72 & \cellcolor{blue!8}90.42 & \cellcolor{blue!8}39.47 \\
            & \cellcolor{blue!8}\hspace{2.5em}Qwen3-VL-8B-Instruct & \cellcolor{blue!8}70.06 & \cellcolor{blue!8}33.55 & \cellcolor{blue!8}45.23 & \cellcolor{blue!8}23.30 & \cellcolor{blue!8}33.27 & \cellcolor{blue!8}5.90 & \cellcolor{blue!8}10.51 & \cellcolor{blue!8}51.66 & \cellcolor{blue!8}71.21 & \cellcolor{blue!8}96.06 & \cellcolor{blue!8}16.90 & \cellcolor{blue!8}37.19 & \cellcolor{blue!8}19.64 & \cellcolor{blue!8}86.46 & \cellcolor{blue!8}33.66 & \cellcolor{blue!8}90.37 & \cellcolor{blue!8}39.83 \\
            \noalign{\vskip 0.25ex}\cdashline{2-19}\noalign{\vskip 0.75ex}
            & Oracle & 70.12 & 38.65 & 50.53 & 25.50 & 34.51 & 10.45 & 15.39 & 52.55 & 74.14 & 97.13 & 17.50 & 38.48 & 21.98 & 87.91 & 35.96 & 91.04 & 42.05 \\
            \bottomrule
        \end{tabular}
    }
    \vspace{0.15in}
    \label{tab:appendix_diverse_lvlms}
\end{table*}

%% file: tables/tab_appendix_granularity.tex
\begin{table}[t]
\centering
\caption{Effect of granularity on the performance for training-based router models. Gn denotes Granularity.}
\renewcommand{\arraystretch}{0.95}
\resizebox{0.5\linewidth}{!}{%
\begin{tabular}{l c c c c}
    \toprule
    & & \multicolumn{2}{c}{\textbf{HotpotQA}} & \textbf{LVBench} \\
    \cmidrule(lr){3-4} \cmidrule{5-5}
    \textbf{Models} & \textbf{Gn} & EM & F1 & Acc \\ 
    \midrule
    \midrule
    \multirow{2}{*}{Qwen3-VL-2B-Instruct} & \redx & 22.25 & 30.38 & 32.05 \\
    & \greencheck & \textbf{26.10} & \textbf{34.61} & \textbf{33.72} \\
    \noalign{\vskip 0.25ex}\cdashline{1-5}\noalign{\vskip 0.75ex}
    \multirow{2}{*}{InternVL3.5-1B} & \redx & 23.00 & 30.89 & 32.05 \\
    & \greencheck & \textbf{25.85} & \textbf{34.29} & \textbf{33.72} \\
    \noalign{\vskip 0.25ex}\cdashline{1-5}\noalign{\vskip 0.75ex}
    \multirow{2}{*}{T5Gemma 2 270M} & \redx & 22.55 & 30.61 & 31.40 \\
    & \greencheck & \textbf{25.90} & \textbf{33.94} & \textbf{33.59} \\
    \bottomrule
\end{tabular}
}
\label{tab:appendix_granularity}
\end{table}

%% file: tables/tab_main_ood.tex
\begin{table*}[t]
    \centering
    \caption{Results of diverse RAG methods on out-of-domain dataset with Qwen3-VL-8B-Instruct across modalities. \textbf{Bold} denotes the best performance and \underline{underlined} indicates the second-best among UniversalRAG variants, using either \textcolor{teal!80}{trained} or \textcolor{blue!80}{training-free} routers. R-L and BERT correspond to ROUGE-L and BERTScore, respectively.}
    \renewcommand{\arraystretch}{0.95}
    \renewcommand{\tabcolsep}{2mm}
    \small
    \resizebox{\textwidth}{!}{
        \begin{tabular}{l c cc cc cc cc c c}
            \toprule
            \multicolumn{1}{c}{} & \textbf{TruthfulQA} & \multicolumn{2}{c}{\textbf{TriviaQA}} & \multicolumn{2}{c}{\textbf{SQuAD}} & \multicolumn{2}{c}{\textbf{2WikiMultiHopQA}} & \multicolumn{2}{c}{\textbf{Visual-RAG}} & \textbf{Cinepile} & \multirow[c]{2}{*}[-0.3em]{\textbf{Avg}} \\
            \cmidrule(lr){2-2} \cmidrule(lr){3-4} \cmidrule(lr){5-6} \cmidrule(lr){7-8} \cmidrule(lr){9-10} \cmidrule(lr){11-11}
            \textbf{Models} & Acc & EM & F1 & EM & F1 & EM & F1 & R-L & BERT & Acc \\
            \midrule
            \midrule
            Naïve & 70.00 & 53.25 & 61.51 & 16.75 & 25.32 & 37.60 & 46.23 & 10.82 & 82.78 & 30.76 & 38.75 \\
            \noalign{\vskip 0.25ex}\cdashline{1-12}\noalign{\vskip 0.75ex}
            ParagraphRAG & 68.86 & 55.82 & 63.78 & 34.40 & 44.27 & 41.35 & 50.86 & 8.95 & 80.91 & 30.42 & 42.62 \\
            DocumentRAG & 68.10 & 52.95 & 61.35 & 18.10 & 27.04 & 48.40 & 58.19 & 8.86 & 80.74 & 30.14 & 41.90 \\
            TableRAG & 66.08 & 51.13 & 59.27 & 9.35 & 16.12 & 33.50 & 44.01 & 8.20 & 80.23 & 29.72 & 37.14 \\
            ImageRAG & 68.48 & 51.89 & 59.74 & 13.90 & 22.65 & 31.15 & 41.86 & 11.64 & 83.36 & 32.71 & 39.18 \\
            ClipRAG & 69.11 & 51.59 & 59.52 & 14.45 & 23.07 & 34.20 & 45.13 & 10.38 & 82.48 & 35.97 & 40.38 \\
            VideoRAG & 68.86 & 51.44 & 59.46 & 14.20 & 22.89 & 33.70 & 44.89 & 10.21 & 82.39 & 37.36 & 40.50 \\
            \noalign{\vskip 0.25ex}\cdashline{1-12}\noalign{\vskip 0.75ex}
            UniRAG & 68.73 & 52.04 & 59.89 & 14.30 & 22.93 & 38.25 & 47.14 & 9.14 & 81.02 & 28.19 & 38.92 \\
            GME & 67.97 & 53.86 & 61.73 & 14.95 & 23.65 & 39.40 & 48.09 & 8.65 & 80.67 & 28.68 & 39.22 \\
            $\text{PE}_{\text{core}}$ & 68.61 & 52.50 & 61.11 & 14.50 & 23.28 & 38.10 & 47.02 & 8.84 & 80.84 & 28.75 & 39.08 \\
            VLM2Vec-V2 & 68.10 & 51.89 & 59.99 & 13.85 & 22.66 & 38.85 & 47.95 & 8.70 & 80.72 & 28.89 & 38.99 \\
            \noalign{\vskip 0.25ex}\cdashline{1-12}\noalign{\vskip 0.75ex}
            MultiRAG & 69.49 & 51.29 & 59.36 & 13.65 & 22.47 & 38.35 & 47.32 & 8.43 & 80.42 & 29.58 & 39.15 \\
            \noalign{\vskip 0.25ex}\cdashline{1-12}\noalign{\vskip 0.75ex}
            \multicolumn{12}{l}{\textbf{UniversalRAG (Ours)}} \\
            \addlinespace[0.2ex]
            \rowcolor{green!8}\multicolumn{12}{l}{\hspace{1em}\textbf{\textit{Trained Routers}}} \\
            \rowcolor{green!8}\hspace{2.5em}Qwen3-VL-2B-Instruct & 69.75 & 54.16 & 62.23 & 31.60 & 41.70 & 45.20 & 54.33 & 10.65 & 82.64 & 33.68 & 44.07 \\
            \rowcolor{green!8}\hspace{2.5em}InternVL3.5-1B & \textbf{69.87} & \textbf{54.46} & \underline{62.45} & 30.75 & 40.97 & 44.85 & 53.89 & 10.88 & 82.79 & 32.64 & 43.80 \\
            \rowcolor{green!8}\hspace{2.5em}T5Gemma 2 270M & 69.24 & 53.71 & 61.90 & 30.60 & 40.84 & 44.70 & 53.74 & 10.52 & 82.58 & 33.19 & 43.61 \\            
            \addlinespace[0.2ex]
            \rowcolor{blue!8}\multicolumn{12}{l}{\hspace{1em}\textbf{\textit{Training-free Routers}}} \\
            \rowcolor{blue!8}\hspace{2.5em}GPT-5 & 69.62 & \textbf{54.46} & \textbf{62.58} & \textbf{31.85} & \textbf{42.02} & \textbf{45.85} & \textbf{54.67} & \underline{11.27} & \underline{83.21} & \textbf{34.10} & \textbf{44.39} \\
            \rowcolor{blue!8}\hspace{2.5em}Qwen3-VL-8B-Instruct & \textbf{69.87} & 54.31 & \underline{62.45} & \underline{31.70} & \underline{41.86} & \underline{45.60} & \underline{54.55} & \textbf{11.33} & \textbf{83.31} & \underline{33.82} & \underline{44.35} \\
            \noalign{\vskip 0.25ex}\cdashline{1-12}\noalign{\vskip 0.75ex}
            Oracle & 70.00 & 55.82 & 63.78 & 34.40 & 44.27 & 48.40 & 58.19 & 11.64 & 83.36 & 37.36 & 46.24 \\
            \bottomrule
        \end{tabular}
    }
    \label{tab:appendix_main_ood}
\end{table*}

%% file: figures/fig_appendix_embedding.tex
\begin{figure*}[t]
    \centering
    \begin{minipage}{0.16\linewidth}
        \centering
        \includegraphics[width=0.97\columnwidth]{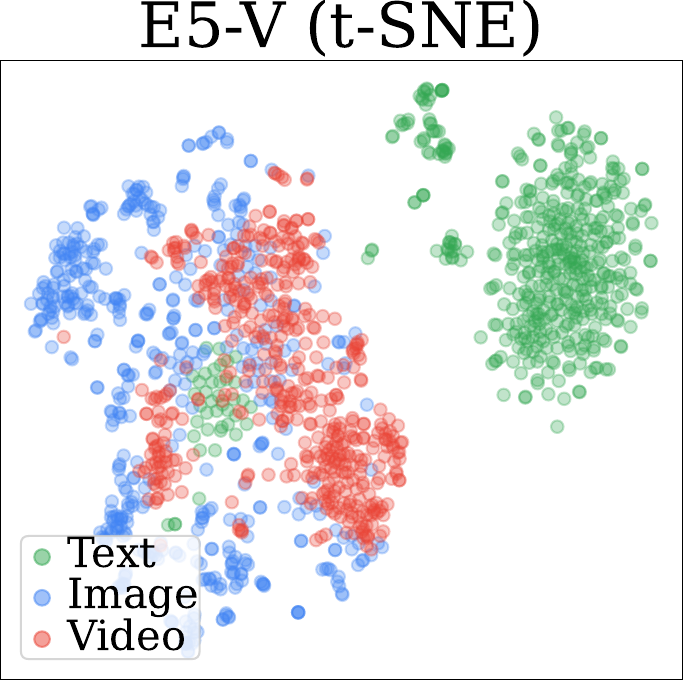} \\
        \vspace{0.1in}
        \includegraphics[width=0.97\columnwidth]{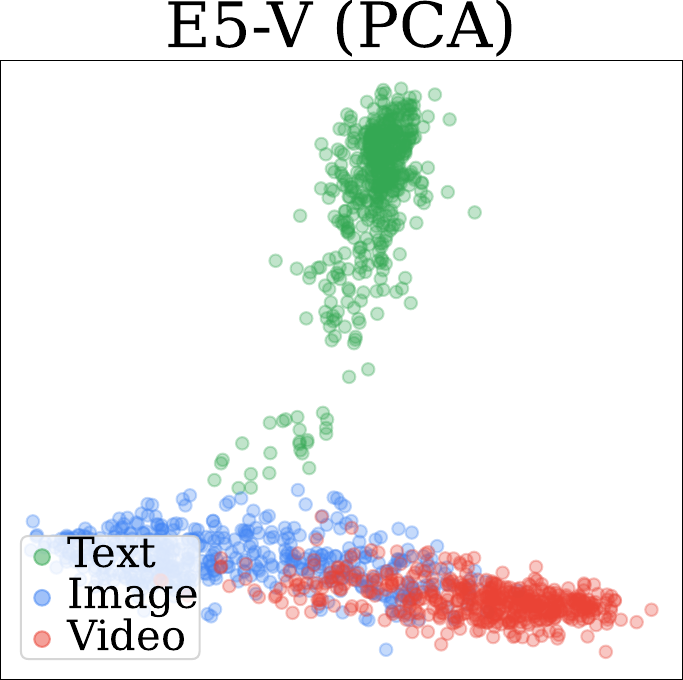}
    \end{minipage}
    \hfill
    \begin{minipage}{0.16\linewidth}
        \centering
        \includegraphics[width=0.97\columnwidth]{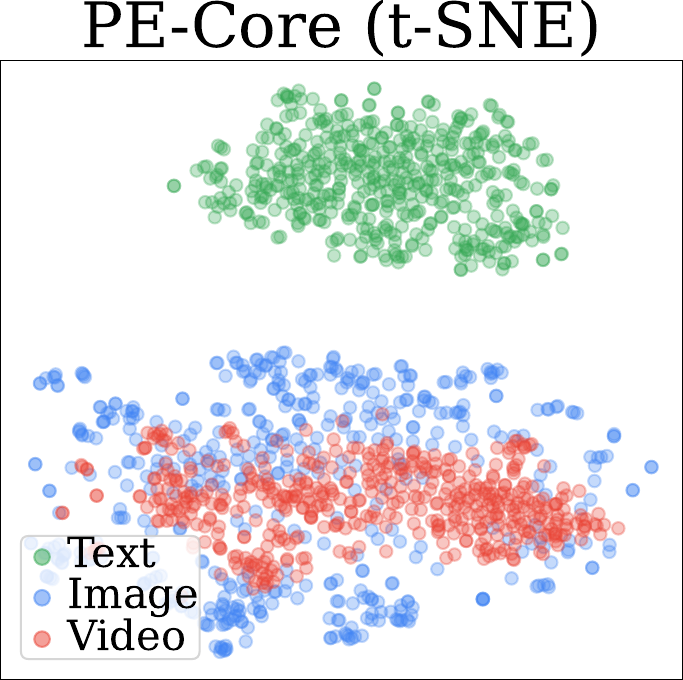} \\
        \vspace{0.1in}
        \includegraphics[width=0.97\columnwidth]{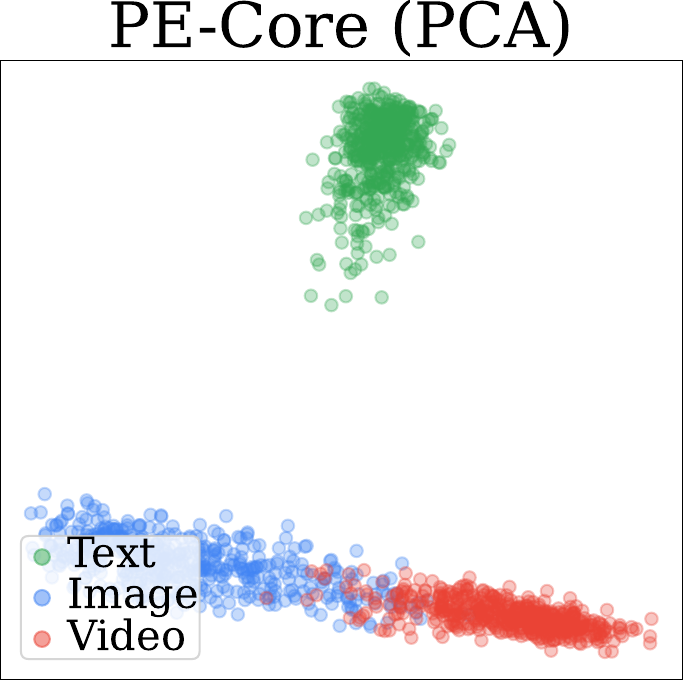}
    \end{minipage}
    \hfill
    \begin{minipage}{0.16\linewidth}
        \centering
        \includegraphics[width=0.97\columnwidth]{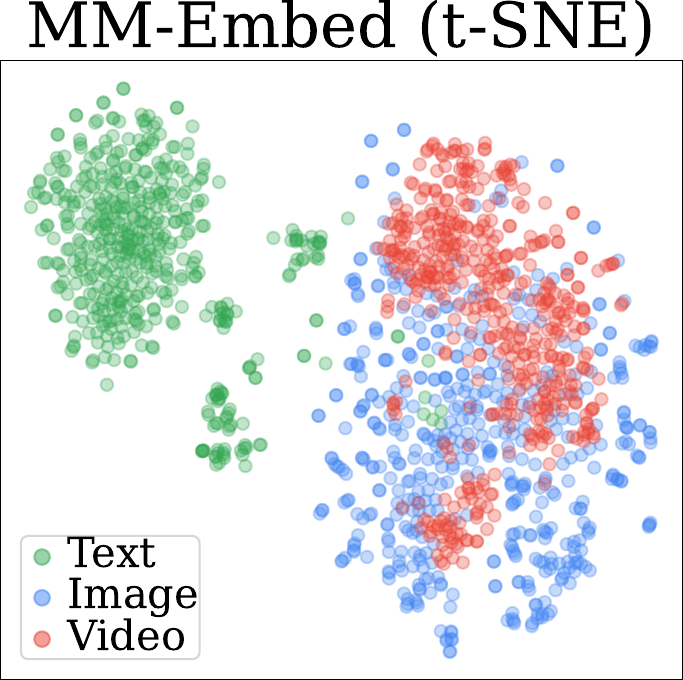} \\
        \vspace{0.1in}
        \includegraphics[width=0.97\columnwidth]{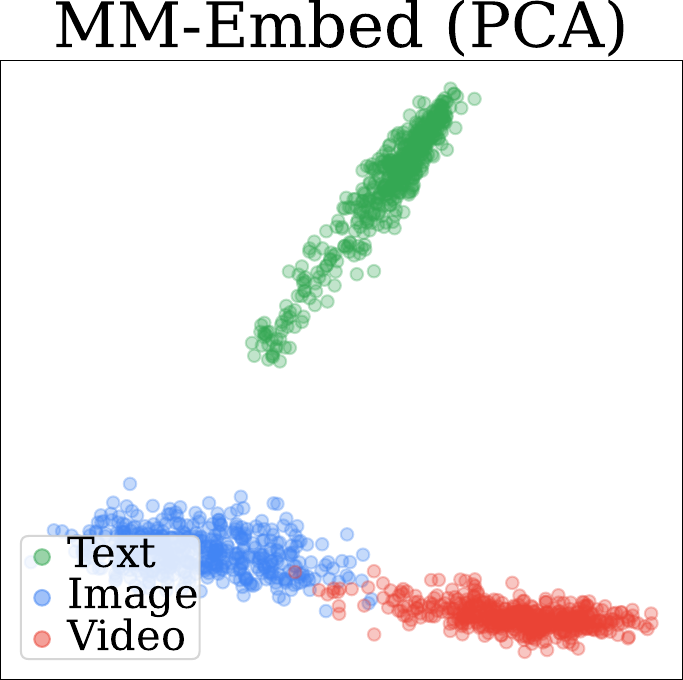}
    \end{minipage}
    \hfill
    \begin{minipage}{0.16\linewidth}
        \centering
        \includegraphics[width=0.97\columnwidth]{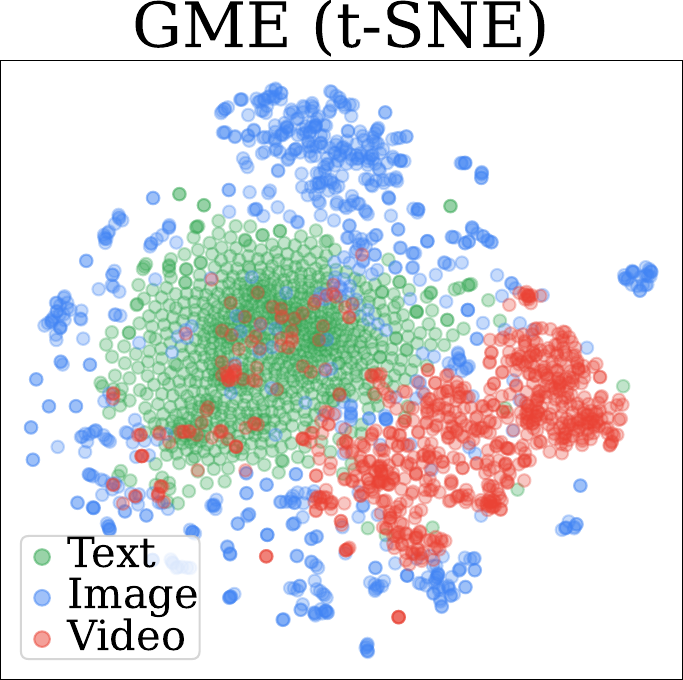} \\
        \vspace{0.1in}
        \includegraphics[width=0.97\columnwidth]{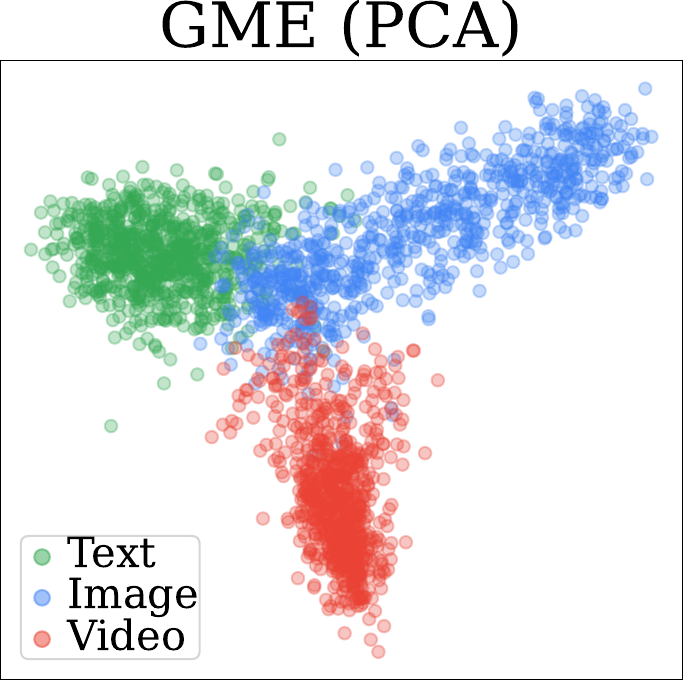}
    \end{minipage}
    \hfill
    \begin{minipage}{0.16\linewidth}
        \centering
        \includegraphics[width=0.97\columnwidth]{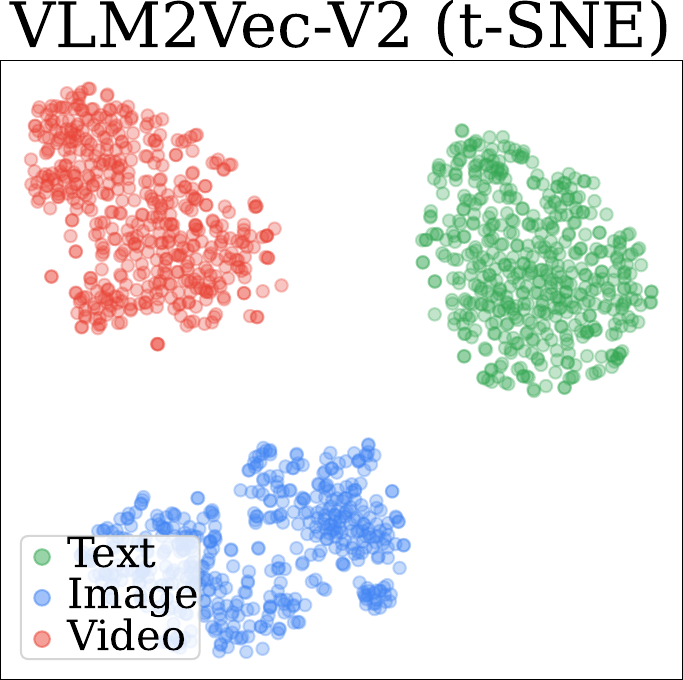} \\
        \vspace{0.1in}
        \includegraphics[width=0.97\columnwidth]{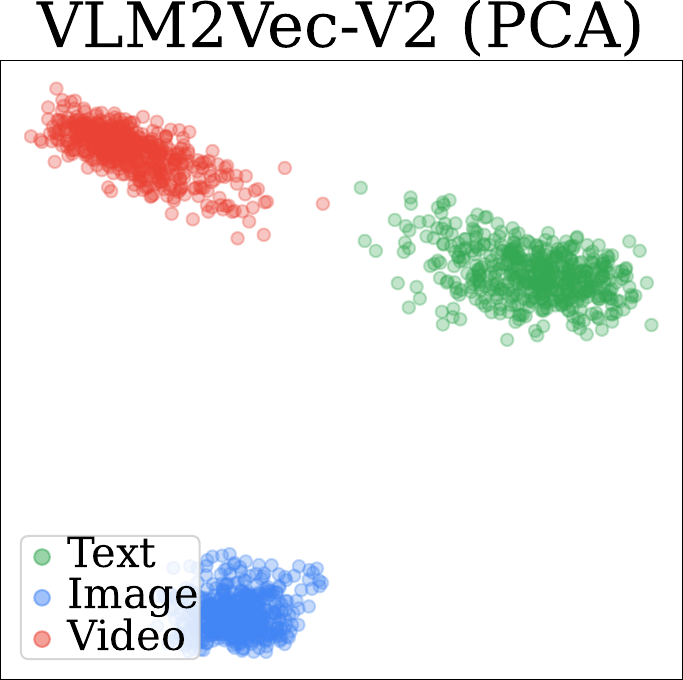}
    \end{minipage}
    \hfill
    \begin{minipage}{0.16\linewidth}
        \centering
        \includegraphics[width=0.97\columnwidth]{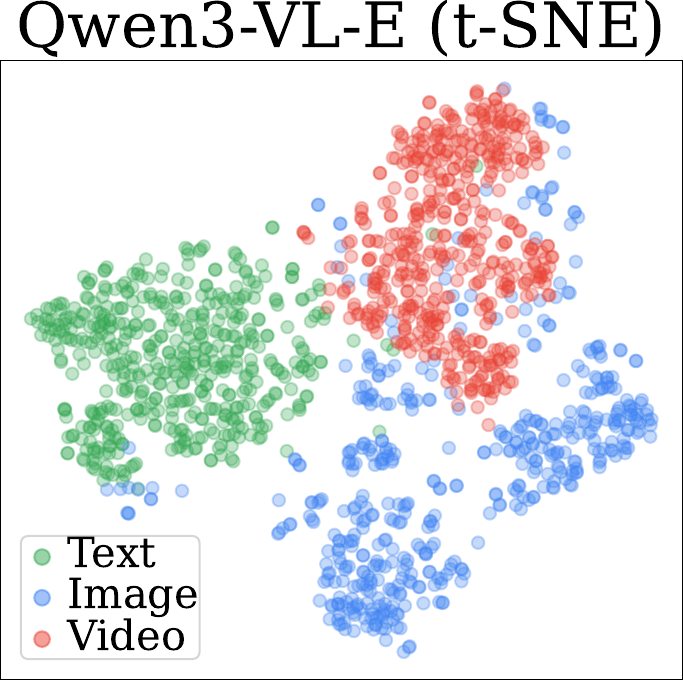} \\
        \vspace{0.1in}
        \includegraphics[width=0.97\columnwidth]{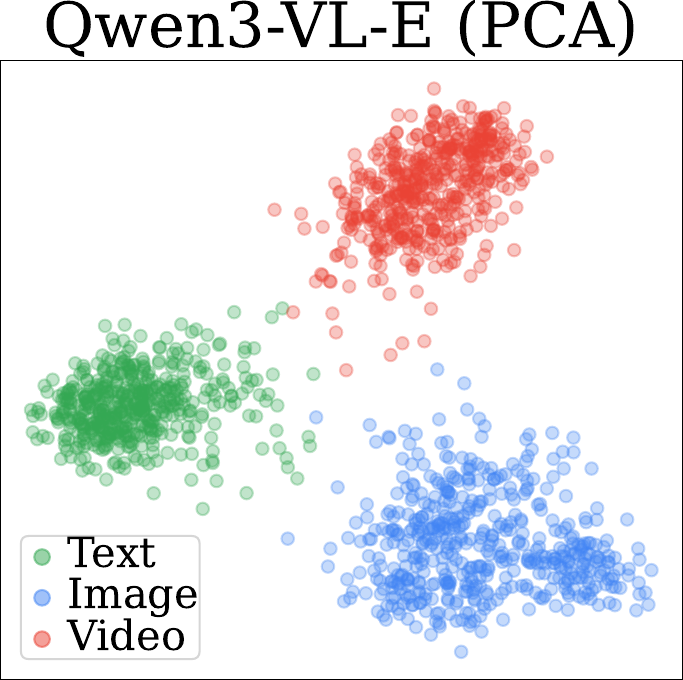}
    \end{minipage}
\caption{Visualization of the unified embedding space across various multimodal encoders.}
\label{fig:appendix_embedding}
\end{figure*}

%% file: tables/tab_qualitative_full.tex
\begin{table*}
    \vspace{-0.2in}
    \centering
    \renewcommand{\arraystretch}{1.0}
    \renewcommand{\tabcolsep}{3mm}
    \caption{Case study comparing RAG on single modality against UniversalRAG (Ours). Traditional RAG models are limited by modality-specific retrieval, often missing relevant knowledge from other modalities. UniversalRAG (Ours) dynamically routes each query to the most appropriate modality, enabling more accurate retrieval and response generation.}
    \small
    \begin{tabular}{l p{0.7\linewidth}}
        \toprule
        \multicolumn{2}{l}{\textbf{Comparison of UniversalRAG to Single-modality RAG}} \\
        \midrule
        \textbf{Question} & What color balloons were displayed on the USNS Carl Brashear during its ship launching ceremony in San Diego? \\
        \midrule
        \textbf{Ground Truth} & The balloons were red, white, and blue that were displayed on the USNS Carl Brashear during the ship launching ceremony in San Diego. \\
        \midrule
        \textbf{No Retrieval} & The USNS Carl Brashear was adorned with blue and gold balloons during its ship launching ceremony in San Diego, reflecting the colors of the U.S. Navy and honoring the legacy of Carl Brashear. \redx \\
        \midrule
        \multirow[t]{2}{*}{\textbf{TextRAG}} & \textbf{Retrieved:} Operation Tomodachi, "Carl Brashear" loaded more than 800 pallets of humanitarian cargo at Sasebo's Juliet pier on March 20 and set sail later that day to join the Navy ships operating off northern Japan. "Brashear" completed 17 underway replenishment missions, delivering more than 1 million gallons (3,800 m³) of fuel to ... \\
        & \vspace{1pt} \textbf{Response:} The documents provided do not mention the color of balloons displayed during the ship launching ceremony of the USNS Carl Brashear in San Diego. \redx \\
        \midrule
        \multirow[t]{2}{*}{\textbf{TableRAG}} & \textbf{Retrieved:} \\
        & \vspace{0.3pt} \resizebox{\linewidth}{!}{%
            \begin{tabular}{c|c|c|c}
                \toprule
                Inmate Name & Register Number & Status & Details \\
                \midrule
                Qian Xuesen & Unlisted† & Held at Terminal Island ... & Chinese-born rocket scientist ... \\
                Liz Renay & Unlisted† & Held at FCI Terminal Island ... & Girlfriend of Los Angeles mob kingpin Mickey Cohen ... \\
                \vdots & \vdots & \vdots & \vdots \\
                \bottomrule
            \end{tabular}
        } \\
        & \vspace{1pt} \textbf{Response:} The color of balloons displayed on the USNS Carl Brashear during its ship launching ceremony in San Diego is not mentioned in the provided documents. \redx \\
        \midrule
        \multirow[t]{2}{*}{\textbf{VideoRAG}} & \textbf{Retrieved:} \\
        & \vspace{1pt}\filmbox{'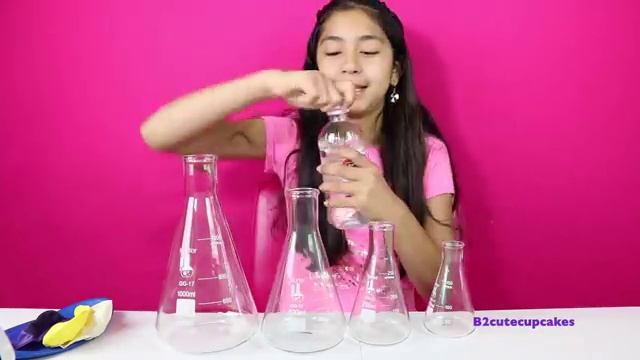'}\filmbox{'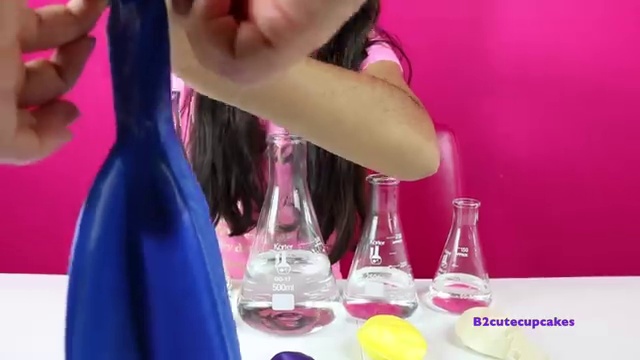'}\filmbox{'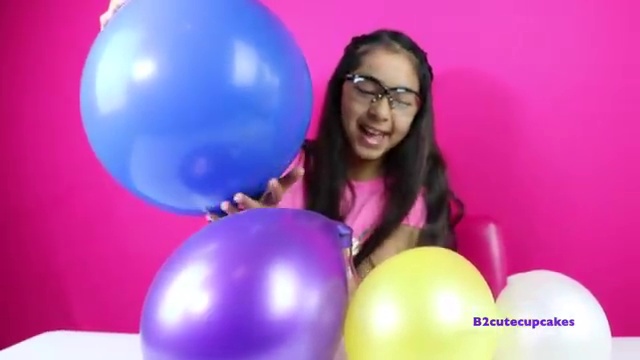'}\filmbox{'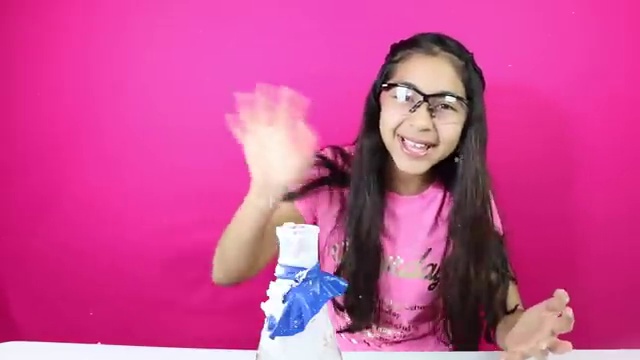'} \\
        & \vspace{1pt} \textbf{Response:} The video does not provide any information about the USNS Carl Brashear or its ship launching ceremony in San Diego, so it is not possible to determine the color of balloons displayed during that event based on the given content. \redx \\
        \midrule
        \multirow[t]{3}{*}{\textbf{UniversalRAG (Ours)}} & \textbf{Routed to:} Image \\
        & \vspace{1pt} \textbf{Retrieved:} \\
        & \includegraphics[width=0.3\linewidth]{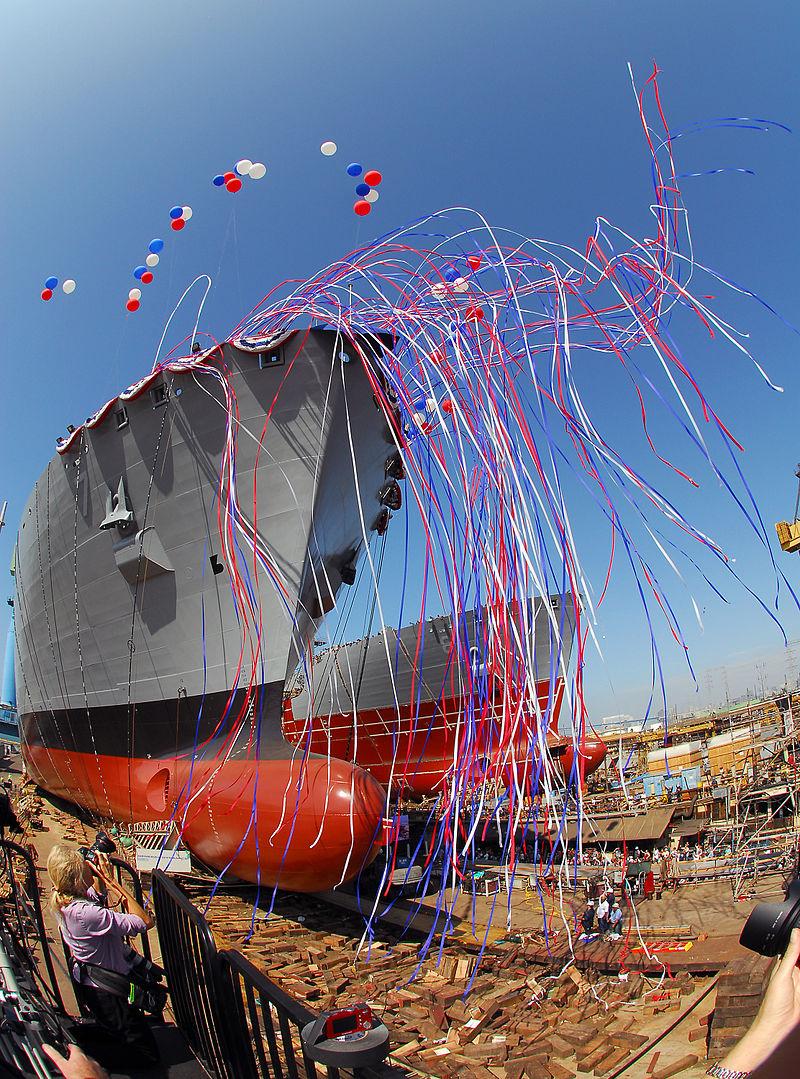} \\
        & \vspace{1pt} \textbf{Response:} During its ship launching ceremony in San Diego, the USNS Carl Brashear was adorned with red, white, and blue balloons. \greencheck \\
        \bottomrule
    \end{tabular}
    \label{tab:qualitative_full}
    \vspace{-0.2in}
\end{table*}

%% file: tables/tab_qualitative_gran_text.tex
\begin{table*}
    \vspace{-0.1in}
    \centering
    \renewcommand{\arraystretch}{1.0}
    \renewcommand{\tabcolsep}{3mm}
    \caption{Case study comparing different levels of text granularity. The user query requires complex retrieval involving multiple entities. ParagraphRAG retrieves limited context centered around a single entity, leading to an incorrect answer. UniversalRAG (Ours) routes the query to the document corpus and retrieves richer document-level information, allowing it to capture both relevant entities and generate the correct response.}
    \small
    \begin{tabular}{l p{0.7\linewidth}}
        \toprule
        \multicolumn{2}{l}{\textbf{Comparison of Using Different Text Granularity}} \\
        \midrule
        \textbf{Question} & George Reed was, along with which \textcolor{blue}{Drake University player}, one of the \textcolor{blue}{greatest running backs in CFL history}? \\
        \midrule
        \textbf{Ground Truth} & Johnny D. Bright \\
        \midrule
        \multirow[t]{2}{*}{\textbf{ParagraphRAG}} & \textbf{Retrieved:} Matthew Reed (born November 30, 1951) is an American former gridiron football player. He played professionally as a quarterback in the World Football League (WFL) and the Canadian Football League (CFL).A graduate of Grambling State University, Reed was a tenth-round selection (240th overall pick) of the Buffalo Bills in the 1973 NFL Draft but did not play in the league. Reed began his pro career with the Birmingham Americans of the WFL. In 1974, he backed up George Mira, completing 77 of 188 passes (41\%) for 1345 yards and 11 touchdowns and 12 interceptions. He was part of their championship team. \\
        & \vspace{1pt} \textbf{Response:} The question contains an error: Matthew Reed is a quarterback, not a running back, and there is no mention of him being associated with Drake University. Additionally, George Reed (a famous CFL running back) is not mentioned in the provided text as being paired with a Drake University player. The text does not contain enough information to answer the question accurately. \redx \\
        \midrule
        \multirow[t]{4}{*}{\textbf{UniversalRAG (Ours)}} & \textbf{Routed to:} Document \\
        & \vspace{1pt} \textbf{Retrieved:} \\
        & \parbox[t]{\linewidth}{ 
        \textsl{George Reed (Canadian football)} George Robert Reed, CM, SOM (born October 2, 1939), is a former American college football and Canadian Football League player. \textcolor{blue}{Reed, along with Mike Pringle and Johnny Bright, is one of the players most often mentioned as being the greatest running back in CFL history}. In November 2006, Reed was voted one of the CFL's Top 50 players (\#2) of the league's modern era by Canadian sports network. \\
        ...\\[0.5ex]
        \textsl{Johnny Bright} Johnny D. Bright (June 11, 1930 – December 14, 1983) was a professional Canadian football player in the Canadian Football League. \textcolor{blue}{He played college football at Drake University.} He is a member of the Canadian Football Hall of Fame, the National Football Foundation's College Football Hall of Fame, the Missouri Valley Conference Hall of Fame, the Edmonton Eskimos Wall of Honour, the Alberta Sports Hall of Fame, and the "Des Moines Register's" Iowa Sports Hall of Fame.
        } \\
        & \vspace{1pt} \textbf{Response:} Johnny Bright \greencheck \\
        \bottomrule
    \end{tabular}
    \label{tab:qualitative_gran_text}
    \vspace{-0.1in}
\end{table*}

%% file: tables/tab_qualitative_gran_video.tex
\begin{table*}
    \centering
    \renewcommand{\arraystretch}{1.0}
    \renewcommand{\tabcolsep}{3mm}
    \caption{Case study comparing different levels of video granularity. The user query requires only a segment of the video to determine the answer. VideoRAG retrieves a broad range of frames across the video, which may include irrelevant content or miss key frames, leading to an incorrect response. UniversalRAG (Ours) routes the query to the clip-level granularity, retrieving more focused and relevant visual information, enabling it to generate the correct response.}
    \small
    \begin{tabular}{l p{0.7\linewidth}}
        \toprule
        \multicolumn{2}{l}{\textbf{Comparison of Using Different Video Granularity}} \\
        \midrule
        \textbf{Question} & What does the protagonist observe through the window after being taken to the utility room in the full episode of Blue Eye Samurai on Netflix?\\&(A) A group of monks sitting cross-legged in the snow\\&(B) A group of citizens chatting together\\&(C) A group of warriors practicing swords\\&(D) A group of samurais eating \\
        \midrule
        \textbf{Ground Truth} & C \\
        \midrule
        \multirow[t]{2}{*}{\textbf{VideoRAG}} & \textbf{Retrieved:} \\
        & \vspace{1pt}\filmbox{'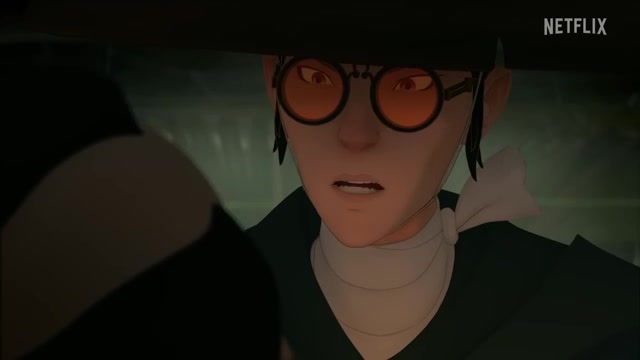'}\filmbox{'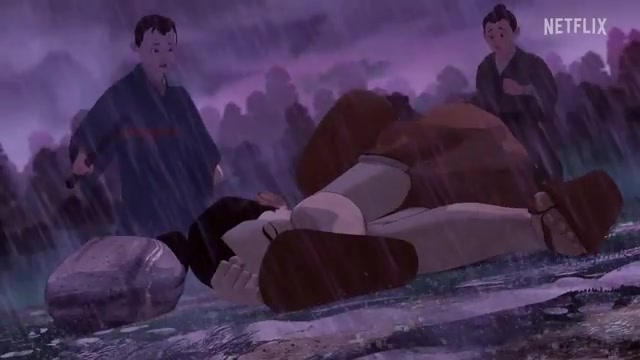'}\filmbox{'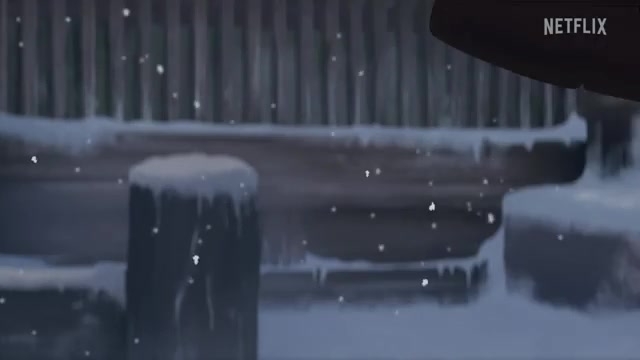'}\filmbox{'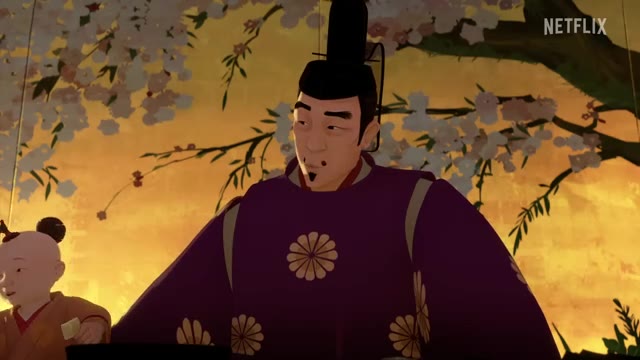'} \\
        & \vspace{1pt}\filmbox{'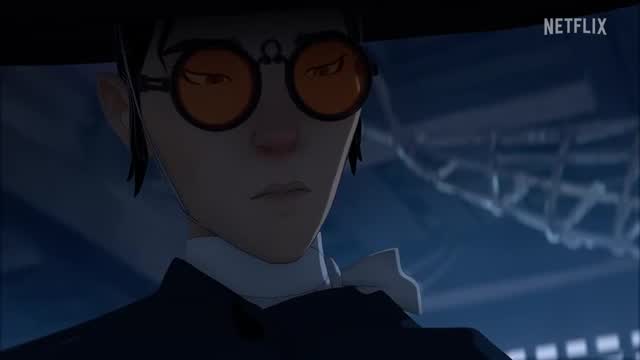'}\filmbox{'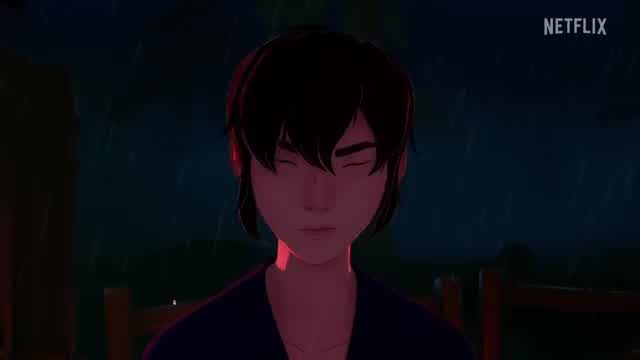'}\filmbox{'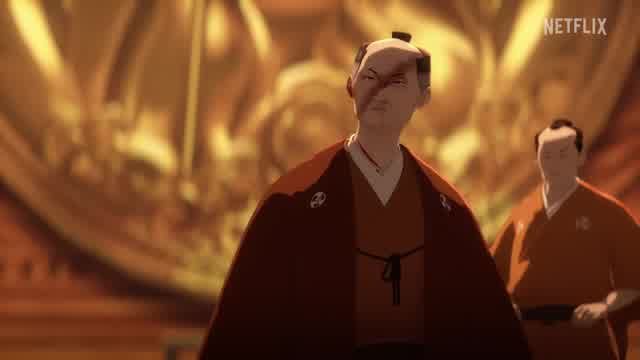'}\filmbox{'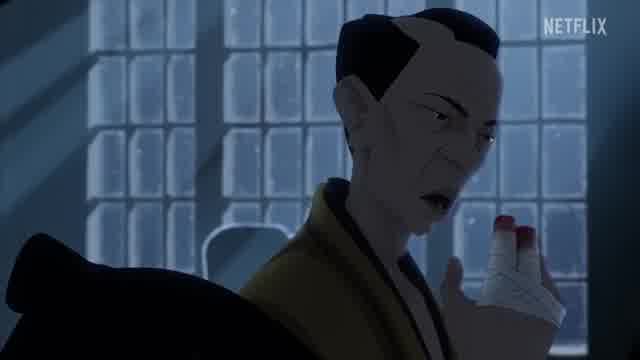'} \\
        & \vspace{1pt}(Timestamp Range: 00:00\textasciitilde1:01:05) \\
        & \vspace{1pt} \textbf{Response:} A \redx \\
        \midrule
        \multirow[t]{3}{*}{\textbf{UniversalRAG (Ours)}} & \textbf{Routed to:} Clip \\
        & \vspace{1pt} \textbf{Retrieved:} \\
        & \vspace{1pt}\filmbox{'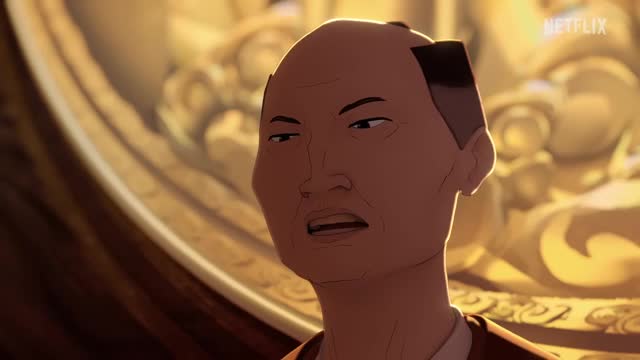'}\filmbox{'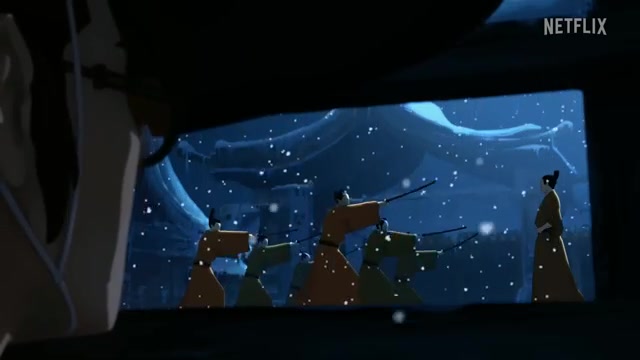'}\filmbox{'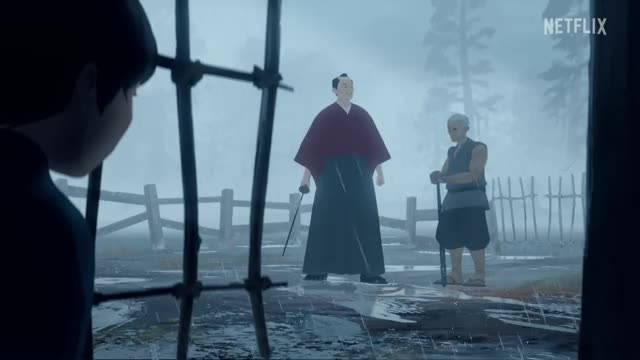'}\filmbox{'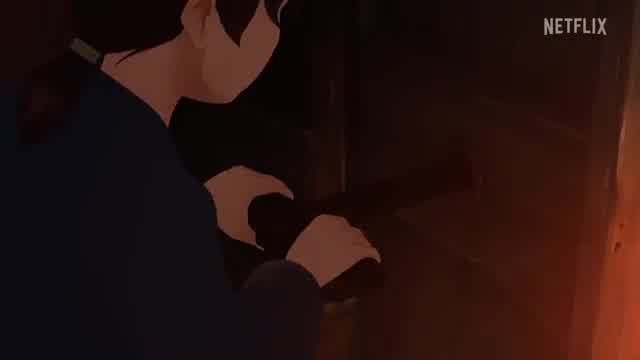'} \\
        & \vspace{1pt}\filmbox{'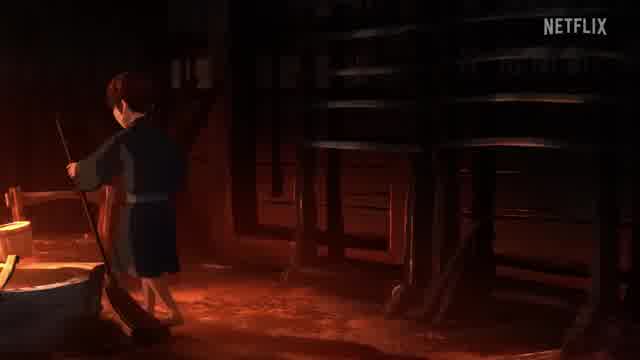'}\filmbox{'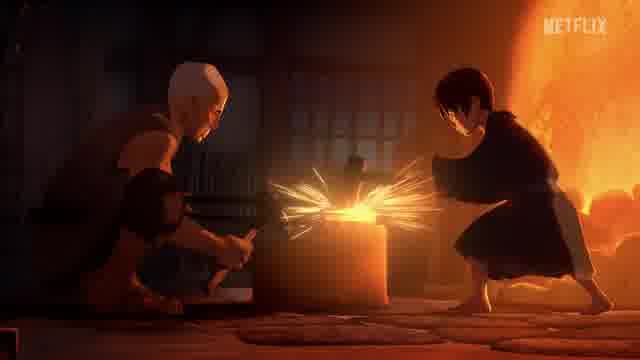'}\filmbox{'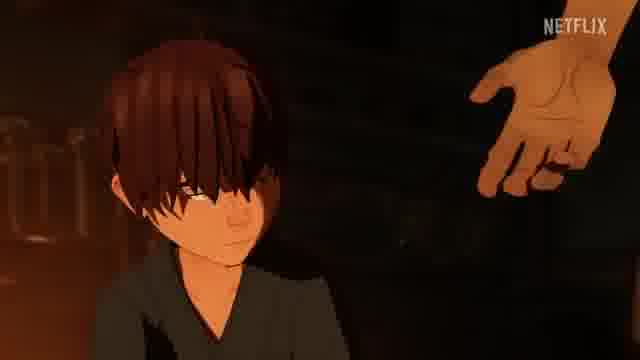'}\filmbox{'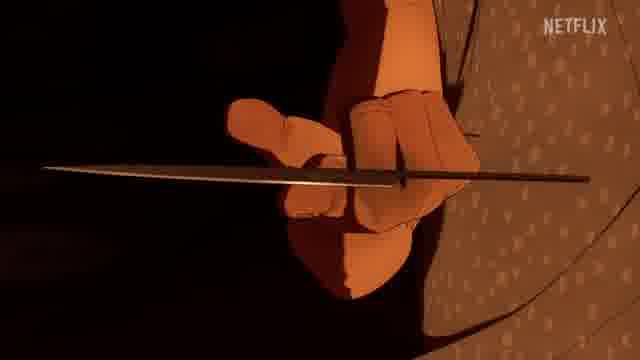'} \\
        & \vspace{1pt}\textcolor{blue}{(Timestamp Range: 33:46\textasciitilde36:56)} \\
        & \vspace{1pt} \textbf{Response:} C \greencheck \\
        \bottomrule
    \end{tabular}
    \label{tab:qualitative_gran_video}
\end{table*}

%% file: tables/tab_qualitative_cross_modal.tex
\begin{table*}
    \vspace{-0.2in}
    \centering
    \renewcommand{\arraystretch}{1.0}
    \renewcommand{\tabcolsep}{3mm}
    \caption{Case study comparing UniversalRAG across uni-modal and cross-modal scenarios. In the uni-modal setup, where only a single prominent modality is used, information can sometimes be incomplete as they require evidence across modalities. UniversalRAG (Ours), with its cross-modal capability, gathers evidence from multiple modalities to generate a more comprehensive response.}
    \small
    \begin{tabular}{l p{0.7\linewidth}}
        \toprule
        \multicolumn{2}{l}{\textbf{Comparison of UniversalRAG across Uni- and Cross-Modal Retrieval}} \\
        \midrule
        \textbf{Question} & What year did an artist known by the mid-1960s in soul and jazz circles for his recording skills release an American record company and label founded by brothers Max and Sol Weiss in 1949? \\
        \midrule
        \textbf{Ground Truth} & 2000 \\
        \midrule
        \multirow[t]{3}{*}{\textbf{UniversalRAG (Uni-modal)}} & \textbf{Routed to:} Paragraph \\
        & \vspace{1pt} \textbf{Retrieved:} \textcolor{blue}{David Axelrod} ( April 17 , 1931 [ nb 1 ] - February 5 , 2017 ) was an American composer , arranger , and producer . After starting out as a staff producer for record companies specializing in jazz , Axelrod became \textcolor{blue}{known by the mid-1960s in soul and jazz circles for his recording skills} . In 1968 , Axelrod embarked on a solo career and released several eccentric albums during the 1970s that showcased his characteristic sound , which combined heavily microphoned drums and baroque orchestration , and avant garde themes ranging from the environment to heightened mental awareness . With his early solo projects , Axelrod was one of the first recording artists to fuse elements of jazz , rock , and R \& B . One of his most important records , Song of Innocence ( 1968 ) , featured instrumental interpretations of 18th-century poet William Blake 's poetry collection of the same name done in a contemporary musical vein , leading one critic at the time to coin the term jazz fusion and numerous hip hop producers to sample the album 's music decades later .\\
        & \vspace{1pt} \textbf{Response:} 1960 \redx \\
        \midrule
        \multirow[t]{3}{*}{\textbf{UniversalRAG (Ours)}} & \textbf{Routed to:} Paragraph+Table \\
        & \vspace{1pt} \textbf{Retrieved:} (Above Paragraph with the following table) \\
        & \vspace{0.3pt} \resizebox{\linewidth}{!}{%
            \begin{tabular}{c|c|c|c|c|c}
                \toprule
                Year & Album & Artist & Genre & Label & Credit \\
                \midrule
                \vdots & \vdots & \vdots & \vdots & \vdots \\
                1998 & Greatest Hits & Joe Cocker & Rock & EMI Electrola & Trombone on You Can Leave Your Hat On \\
                \textcolor{blue}{2000} & The Axelrod Chronicles & \textcolor{blue}{David Axelrod} & Jazz , funk , soul & Fantasy & Trombone \\
                2004 & Ultimate Collection & Joe Cocker & Rock & Hip-O , A \& M & Horn on You Can Leave Your Hat On \\
                \vdots & \vdots & \vdots & \vdots & \vdots \\
                \bottomrule
            \end{tabular}
        } \\
        & \vspace{1pt} \textbf{Response:} 2000 \greencheck \\
        \bottomrule
    \end{tabular}
    \label{tab:qualitative_cross}
    \vspace{-0.2in}
\end{table*}

%% file: tables/tab_qualitative_failure.tex
\begin{table*}
    \centering
    \renewcommand{\arraystretch}{1.0}
    \renewcommand{\tabcolsep}{3mm}
    \caption{Failure cases in modality routing with UniversalRAG (Ours).}
    \small
    \begin{tabular}{m{0.5\linewidth} c c}
        \toprule
        \textbf{Question} & \textbf{Ground Truth} & \textbf{UniversalRAG (Ours)} \\
        \midrule
        \midrule
        What language does the French word polytechnique come from? & Paragraph & No \\
        \midrule
        Who is seated to the right of Kobe in the Jimmy Kimmel tribute show? & Clip & Image \\
        \midrule
        Which book by William A. Dembski summarizes the concepts he introduced about intelligent design in another of his works? & Document & Paragraph \\
        \midrule
        What is the main cause of Lee Chong Wei losing points in the first half of his semi-final match against Lin Dan in the Rio 2016 Olympics replay? & Video & Clip \\
        \midrule
        What is at the top of Hanbit Tower at Expo Science Park? & Paragraph+Image & Paragraph \\
        \bottomrule
    \end{tabular}
    \label{tab:qualitative_failure}
\end{table*}

%% file: figures/fig_prompts.tex
\begin{figure*}
    \centering
    \vspace{-1cm}
    \begin{tcolorbox}[colback=gray!5!white, colframe=black!75!white]
        Classify the following query into one or more categories from: \textbf{[No, Paragraph, Document, Table, Image, Clip, Video]}, based on whether it requires retrieval-augmented generation (RAG) and the most appropriate modality. Consider:

        \vspace{0.3cm}

        \begin{itemize}[topsep=0pt, itemsep=0ex, parsep=0pt, left=10pt]
            \item \textbf{No}: The query can be answered directly with common knowledge, reasoning, or computation without external data.
            \item \textbf{Paragraph}: The query requires retrieving factual descriptions, straightforward explanations, or concise summaries from a single source.
            \item \textbf{Document}: The query requires multi-hop reasoning, combining information from multiple sources or documents to form a complete answer.
            \item \textbf{Table}: The query requires information that is best represented in a tabular format, often involving comparisons or structured data.
            \item \textbf{Image}: The query focuses on visual aspects like appearances, structures, or spatial relationships.
            \item \textbf{Clip}: The query targets a short, specific moment or event within a video, without needing full context.
            \item \textbf{Video}: The query requires understanding dynamic events, motion, or sequences over time in a video.
        \end{itemize}

        \vspace{0.3cm}

        \textbf{Examples:}
        \begin{itemize}[topsep=0pt, itemsep=0ex, parsep=0pt, left=10pt]
            \item "What is the capital of France?" $\rightarrow$ \textbf{No}
            \item "What is the birth date of Alan Turing?" $\rightarrow$ \textbf{Paragraph}
            \item "Which academic discipline do computer scientist Alan Turing and mathematician John von Neumann have in common?" $\rightarrow$ \textbf{Document}
            \item "Among the recipients of the Turing Award, who had the earliest birth year?" $\rightarrow$ \textbf{Table}
            \item "Describe the appearance of a blue whale." $\rightarrow$ \textbf{Image}
            \item "Describe the moment Messi scored his goal in the 2022 World Cup final." $\rightarrow$ \textbf{Clip}
            \item "Explain how Messi scored his goal in the 2022 World Cup final." $\rightarrow$ \textbf{Video}
            \item "Solve 12 × 8." $\rightarrow$ \textbf{No}
            \item "Who played a key role in the development of the iPhone?" $\rightarrow$ \textbf{Paragraph}
            \item "Which Harvard University graduate played a key role in the development of the iPhone?" $\rightarrow$ \textbf{Document}
            \item "What is the cheapest iPhone model available in 2023?" $\rightarrow$ \textbf{Table}
            \item "Describe the structure of the Eiffel Tower." $\rightarrow$ \textbf{Image}
            \item "Describe the moment Darth Vader reveals he is Luke's father in Star Wars." $\rightarrow$ \textbf{Clip}
            \item "Analyze the sequence of events leading to the fall of the Empire in Star Wars." $\rightarrow$ \textbf{Video}
            \item "Describe the visual appearance and habitat of the blue whale." $\rightarrow$ \textbf{Paragraph+Image}
            \item "Compare the architectural features shown in Gothic and Renaissance cathedrals." $\rightarrow$ \textbf{Image+Table}
            \item "Describe the moment of the moon landing and explain the mission details." $\rightarrow$ \textbf{Paragraph+Clip}
        \end{itemize}

        \vspace{0.3cm}

        Classify the following query: \{\texttt{query}\} \\
        Provide only the category or categories combined with `+'.
    \end{tcolorbox}
    \caption{Prompt for query routing in a training-free manner. The prompt defines each category with concise criteria and illustrative examples. Specifically, examples are designed to contrast closely related cases: for example, Paragraph vs. Document for simple fact retrieval vs. multi-hop reasoning; and Clip vs. Video for short specific moments vs. long-term sequential understanding, highlighting the key aspect that differentiates each category.}
    \vspace{-1cm}
    \label{fig:prompt_route}
\end{figure*}

\begin{figure*}
    \centering
    \begin{tcolorbox}[colback=gray!5!white, colframe=black!75!white]
        Classify the following query into one or more categories from: \textbf{[No, Paragraph, Passage, Section, Document, ... , Clip, Sequence, Segment, Video]}, based on whether it requires retrieval-augmented generation (RAG) and the most appropriate modality. Consider:

        \vspace{0.3cm}

        \begin{itemize}[topsep=0pt, itemsep=0ex, parsep=0pt, left=10pt]
            \item \textbf{Paragraph}: The query requires retrieving factual descriptions, straightforward explanations, or concise summaries from a single source.
            \item \textbf{Passage}: The query requires a detailed block of text (a few paragraphs) from a single source, with added context.
            \item \textbf{Section}: The query requires retrieving an extensive section of a document explaining a sub-topic, possibly with examples or elaboration.
            \item \textbf{Document}: The query requires multi-hop reasoning, combining information from multiple sources or documents to form a complete answer.
            \item \textbf{Clip}: The query targets a short, specific moment or event within a video, without needing full context.
            \item \textbf{Sequence}: The query targets a continuous stretch of related shots (about 10 minutes) that together form a self-contained mini-narrative or process, providing more context and flow than a standalone clip.
            \item \textbf{Segment}: The query targets a longer portion of a video (about 30 minutes) capturing a meaningful sub-scene or subplot-rich and cohesive enough to serve as its own chapter-like unit.
            \item \textbf{Video}: The query requires understanding dynamic events, motion, or sequences over time in a video.
        \end{itemize}

        \vspace{0.3cm}

        \textbf{Examples:}
        \begin{itemize}[topsep=0pt, itemsep=0ex, parsep=0pt, left=10pt]
            \item "What is the birth date of Alan Turing?" $\rightarrow$ \textbf{Paragraph}
            \item "Summarize Alan Turing's concept of the Turing Machine." $\rightarrow$ \textbf{Passage}
            \item "Explain Alan Turing's contributions to cryptography during WWII." $\rightarrow$ \textbf{Section}
            \item "Which academic discipline do computer scientist Alan Turing and mathematician John von Neumann have in common?" $\rightarrow$ \textbf{Document}
            \item "Describe the moment Messi scored his goal in the 2022 World Cup final." $\rightarrow$ \textbf{Clip}
            \item "Detail the sequence of passes and movements leading to Messi's goal in the 2022 World Cup final." $\rightarrow$ \textbf{Sequence}
            \item "Describe the build-up sequence during the mid-game period of the 2022 World Cup final." $\rightarrow$ \textbf{Segment}
            \item "Analyze how Argentina won the 2022 World Cup." $\rightarrow$ \textbf{Video}
        \end{itemize}

        \vspace{0.3cm}

        Classify the following query: \{\texttt{query}\} \\
        Provide only the category or categories combined with `+'.
    \end{tcolorbox}
    \caption{Prompt for query routing in a training-free manner with additional granularity choices. Only the components that differ from \cref{fig:prompt_route} are shown, including the task objective and few-shot examples.}
    \label{fig:prompt_route_gran}
\end{figure*}

\begin{figure*}
    \begin{tcolorbox}[colback=gray!5!white, colframe=black!75!white]
        Evaluate whether the query can be answered using general knowledge about the image's subject rather than relying solely on details unique to the provided image, and verify that the answer is obtainable from the image and the query.
    
        \begin{itemize}[topsep=0pt, itemsep=0ex, parsep=0pt, left=10pt]
            \item Respond "yes" if:
            \begin{enumerate}[topsep=0pt, itemsep=0ex, parsep=0pt]
                \item The query can be fully answered using general knowledge about the subject.
                \item The answer can be derived solely from the image and the query, without needing image-specific details.
            \end{enumerate}
            \item Respond "no" if either condition is not met.
        \end{itemize}
        
        \vspace{0.3cm}
        
        \textbf{Example 1:}
        \begin{itemize}[topsep=0pt, itemsep=0ex, parsep=0pt, left=10pt]
            \item Image: A portrait of Donald Trump
            \item Query: What is the color of Trump's hair?
            \item Answer: White
            \item Response: "yes"
        \end{itemize}
        
        \vspace{0.3cm}
        
        \textbf{Example 2:}
        \begin{itemize}[topsep=0pt, itemsep=0ex, parsep=0pt, left=10pt]
            \item Image: A close-up photo of a light bulb
            \item Query: What is the color of the light bulb in this image?
            \item Answer: Yellow
            \item Response: "no"
        \end{itemize}
    \end{tcolorbox}
    \caption{Prompt to filter queries for WebQA.}
    \label{fig:prompt_filter}
\end{figure*}

\begin{figure*}[htbp]
    \centering
    \begin{tcolorbox}[colback=gray!5!white, colframe=black!75!white]
        You will receive a query from a video QA dataset and the title of the corresponding video on YouTube. I want you to paraphrase the query by replacing "in the video?", "of the video", or similar phrases with references to the video content naturally. The output should sound as if a human is asking ChatGPT, and should not explicitly mention the exact name of the video or even parts of the title. However, the rephrased query should contain enough implicit information about the video to allow the model to identify it. Try to reduce the chance of the model getting confused between multiple possible video candidates. If there could be multiple video matches for a given query, try to include more information in the rephrased query. \\

        \textbf{Example 1:}
        \begin{itemize}[topsep=0pt, itemsep=0ex, parsep=0pt, left=10pt]
            \item Query: What year appears in the opening caption of the video?
            \item Video Title: Blue Eye Samurai | Hammerscale | Full Episode | Netflix
            \item Upload Date: 2023-11-05
            \item Channel Name: Netflix
            \item Rephrased Output: What year appears in the opening caption of the Blue Eye Samurai episode on Netflix?
        \end{itemize}

        \vspace{0.3cm}

        \textbf{Example 2:}
        \begin{itemize}[topsep=0pt, itemsep=0ex, parsep=0pt, left=10pt]
            \item Query: After the vlogger sees a dog with an advertisement from the company named Smitten, camera changes to the scene with \_\_\_.
            \item Video Title: My ICELAND Experience | Ultimate Travel Vlog
            \item Upload Date: 2022-10-26
            \item Channel Name: Kallmekris
            \item Rephrased Output: After spotting a dog with a Smitten advertisement, what scene does the camera transition to in Kallmekris's Iceland travel vlog from 2022?
        \end{itemize}
    \end{tcolorbox}
    \caption{Prompt to rephrase queries using video metadata for LVBench and CinePile.}
    \label{fig:prompt_rephrase}
\end{figure*}